\documentclass[10pt]{article}

\usepackage{gta_arxiv}

\usepackage{amsmath,amsfonts,bm}

\def\eqref#1{equation~\ref{#1}}

\def\1{\bm{1}}

\DeclareMathAlphabet{\mathsfit}{\encodingdefault}{\sfdefault}{m}{sl}
\SetMathAlphabet{\mathsfit}{bold}{\encodingdefault}{\sfdefault}{bx}{n}

\usepackage{microtype}
\usepackage[utf8]{inputenc} 
\usepackage{graphicx}
\usepackage{flafter}
\usepackage{booktabs}       
\usepackage{amsmath}
\usepackage{amssymb}
\usepackage{amsfonts}       
\usepackage{nicefrac}       
\usepackage[table]{xcolor}  
\usepackage{enumitem}
\usepackage{makecell}
\usepackage{multirow}
\usepackage{wrapfig}
\usepackage{needspace}
\usepackage{array}
\usepackage{tabularx}
\usepackage{tcolorbox}
\usepackage{listings}
\usepackage{url}
\usepackage{hyperref}

\definecolor{darkblue}{rgb}{0, 0, 0.5}
\hypersetup{colorlinks=true, citecolor=darkblue, linkcolor=darkblue, urlcolor=darkblue}

\title{GTA: Graph Theory Agent and Benchmark for Algorithmic Graph Reasoning with LLMs}

\newcommand{\AuthorNames}{Zixiang Xu, Yanbo Wang, Chenxi Wang, Lang Gao, Zirui Song, Yue Huang, Zhaorun Chen, Xiangliang Zhang, Xiuying Chen}
\author{%
  \name Zixiang Xu\textsuperscript{1}, Yanbo Wang\textsuperscript{2}, Chenxi Wang\textsuperscript{3}, Lang Gao\textsuperscript{3}, Zirui Song\textsuperscript{3} \\
  \name Yue Huang\textsuperscript{4}, Zhaorun Chen\textsuperscript{5}, Xiangliang Zhang\textsuperscript{4}, Xiuying Chen\textsuperscript{3}\thanks{Corresponding author: Xiuying Chen (\href{mailto:xiuying.chen@mbzuai.ac.ae}{\texttt{xiuying.chen@mbzuai.ac.ae}}).} \\[0.5em]
  \addr \textsuperscript{1}University of Southern California \quad \textsuperscript{2}University of California, Los Angeles \\
  \addr \textsuperscript{3}Mohamed bin Zayed University of Artificial Intelligence \\
  \addr \textsuperscript{4}University of Notre Dame \quad \textsuperscript{5}University of Chicago
}

\date{}

\hypersetup{
  pdftitle={GTA: Graph Theory Agent and Benchmark for Algorithmic Graph Reasoning with LLMs},
  pdfauthor={\AuthorNames},
  pdfsubject={Algorithmic graph reasoning with large language models},
  pdfkeywords={graph reasoning, large language models, benchmark, adaptive representation}
}

\begin{document}

\maketitle
\begin{abstract}
Large Language Models (LLMs) are increasingly asked to reason over structured data such as graphs, yet how reliably they can carry out multi-step graph algorithms in language remains unclear. Existing evaluations tend to use simple tasks on small graphs, to score code generation rather than reasoning over the graph itself, or to fix a single input format. We introduce \textbf{Graph Theory Bench (GT Bench)}, a benchmark covering 24 classical graph problems in 44 task--structure settings, with over 100{,}000 examples across four representations: natural language, structured language, adjacency list, and adjacency matrix. Evaluating eight LLMs on GT Bench shows that accuracy is strongly tied to the input representation, that the best representation shifts with graph density, size, and topology as well as with the model, and that this sensitivity persists, attenuated, in the strongest reasoning models. Building on these observations, we propose the \textbf{Graph Theory Agent (GTA)}, which pairs a preference-trained representation selector with plan-and-decompose scaffolding around a frozen executor LLM. GTA lifts Phi-4 from 53.5\% to 69.1\% on the benchmark's easy split and from 33.0\% to 41.5\% on its hard split, outperforming eight prompting and agent baselines, and transfers without retraining to GraCoRe and NLGraph. Code for benchmark generation and evaluation: \url{https://github.com/xzx34/GTA}. The project homepage is available at \url{https://xzx34.github.io/gta/}.
\end{abstract}

\section{Introduction}
Large Language Models (LLMs) have moved well beyond text processing into social-science applications \citep{huang2024creation,xu2026socialmaze}, medicine \citep{zhou2024pre,tian2024opportunities,liu2023deid,chen2025evaluating}, and robotics~\citep{song2024hazards}. As their reach widens, they are increasingly expected to reason about structured data, including social networks~\citep{wang2025decoding}, interaction graphs for session search~\citep{wu2024unify}, and molecular structures~\citep{ross2022molecular}. These applications connect entities through relations that can be represented as vertices and edges. Working with such data takes more than recognizing that a graph is present; the model must interpret its structure and sustain multi-step algorithmic reasoning over it.

Existing graph benchmarks evaluate several distinct capabilities. GraphEval36K \citep{wu2025grapheval36kbenchmarkingcodingreasoning} measures graph problem solving through generated code, whereas NLGraph \citep{wang2023can} and GraCoRe \citep{yuan2024gracore} evaluate answers to structural and relational questions over graph inputs. GRBench \citep{jin2024graph} evaluates question answering with knowledge from text-attributed graphs. Representation studies further show that graph encoding can change LLM performance \citep{fatemi2023talk}. These complementary settings motivate a controlled evaluation that combines diverse classical graph problems, equivalent input encodings, and language-only execution.

To support this evaluation, we introduce \textbf{Graph Theory Bench (GT Bench)}, a benchmark specifically designed to assess the multi-step algorithmic reasoning capabilities of LLMs on graph-theoretic problems. GT Bench covers 24 classical graph problems in 44 task--structure settings, totaling more than 100{,}000 examples across four representations. The examples are stratified into \textit{easy} and \textit{hard} subsets covering tasks from basic connectivity checks to complex problems like minimum-cost maximum-flow (MCMF). Each instance is presented in four input modalities (natural language, structured language, adjacency list, adjacency matrix) to test model robustness and the impact of representation on reasoning. A parameterized generation pipeline supports flexible control of graph families (e.g., weighted/unweighted, sparse/dense, tree-structured) and sizes, with node and edge counts calibrated to contemporary LLM context windows: large enough to require genuine multi-step reasoning yet compact enough to fit reliably within prompts. Unlike benchmarks focusing on code generation or rudimentary queries, GT Bench targets tasks requiring genuine algorithmic thinking and combinatorial reasoning. The accompanying repository contains sample data and code for graph generation and evaluation.

Our experiments on GT Bench expose a strong dependency between input representation and LLM performance. On a single task the choice of format can move accuracy by close to thirty points (Connectivity on sparse graphs spans 60.5\% to 89.1\% across the four formats), and the winning format varies systematically with graph structure: adjacency matrices excel on dense graphs, natural language on sparse graphs, and structured language on trees. Building on this finding, we propose the \textbf{Graph Theory Agent (GTA)}, a framework that improves LLM graph reasoning through adaptive input representation selection and systematic decomposition of the solution into manageable sub-steps. GTA operates under a \emph{frozen executor} paradigm, keeping the base LLM unmodified while optimizing lightweight auxiliary components for representation selection and algorithmic decomposition.

Our focus on language-only reasoning is deliberate: it measures whether a model can interpret a graph, maintain intermediate algorithmic state, and carry out a sequence of dependent operations. GT Bench includes polynomial-time problems such as shortest path and minimum-cost maximum-flow, as well as NP-hard optimization problems such as Maximum Clique and Maximum Independent Set and NP-complete decision problems such as Hamiltonian Path and Circuit. Evaluating these tasks through language exposes the reasoning steps performed by the model itself. Code-augmented evaluation measures an additional capability: translating a problem into an executable algorithm and using its output.

Our main contributions are: (1) GT Bench, a benchmark for multi-step algorithmic reasoning of LLMs on graph-theoretic problems, with four equivalent graph representations; (2) an empirical characterization of representation sensitivity, showing how the best format depends on task, graph structure, and model, and that the effect survives in strong reasoning models; and (3) GTA, an agent framework that adaptively selects representations and decomposes problem solving around a frozen executor, improving accuracy across benchmarks and executor models.

\section{Related Work}
\label{sec:related_work}

\subsection{LLMs and Graph Reasoning Benchmarks}
\label{appendix:lagrb}
Graph benchmarks differ in their inputs and execution protocols. GraphEval36K \citep{wu2025grapheval36kbenchmarkingcodingreasoning} evaluates generated programs on graph problems. NLGraph \citep{wang2023can} covers eight tasks, including maximum flow and Hamiltonian path; GPT4Graph \citep{guo2023gpt4graph} evaluates structural and semantic understanding; and GraCoRe \citep{yuan2024gracore} covers 19 tasks on pure and heterogeneous graphs. GRBench \citep{jin2024graph} focuses on question answering over text-attributed graphs, while ProofWriter \citep{tafjord2020proofwriter} tests rule-based inference in natural language. HGB \citep{li2023hybrid} provides hybrid-graph datasets for graph neural networks, and HiGPT \citep{tang2024higpt} addresses heterogeneous graph learning. VisionGraph \citep{li2024visiongraph} evaluates multi-step graph-theoretic reasoning from visual inputs. GraphArena \citep{tang2025grapharena} covers ten polynomial-time and NP-complete tasks and distinguishes correct, suboptimal, hallucinatory, and missing answers. ProGraph \citep{li2024prograph} evaluates library-calling programs; GrAlgoBench \citep{zhang2026gralgobench} reports accuracy below 50\% beyond 120 nodes in its evaluation of large reasoning models; LLM4Hypergraph \citep{feng2024hypergraph} studies seven hypergraph languages; and LLM4DyG \citep{zhang2024llm4dyg} probes spatial-temporal reasoning on dynamic graphs. GT Bench contributes a controlled comparison of four standard encodings across 44 task--structure settings for language-only algorithmic reasoning.

A closely related and fast-growing thread studies how graph \emph{representation} affects LLM performance. Format sensitivity is not unique to graphs: \citet{sclar2024quantifying} show that semantically equivalent changes to prompt formatting can swing accuracy by tens of points on general tasks, and the effect survives scale and instruction tuning. Graphs make the phenomenon acute, because the same structure admits several equally standard encodings. ``Talk Like a Graph'' \citep{fatemi2023talk} demonstrated that the choice among them matters, and a series of recent studies has sharpened this picture. \citet{firooz2024lostindistance} show that accuracy drops as related facts are placed farther apart in the serialized context; GraphSOS \citep{chu2025graphsos} finds that merely shuffling node and edge order can swing performance between strong and near-random, and learns to select serialization orders and sample informative subgraphs; \citet{ge2025can} similarly show that the order in which edges are described shifts accuracy in task-dependent ways; and \citet{herbst2025serialization} demonstrate that LLM graph reasoners are not invariant to node relabeling, edge reordering, or syntax, and that fine-tuning can even amplify this sensitivity. Others design new encodings, such as the interpretable color-based scheme of \citet{zangari2026colorful}, or compare textualization strategies for knowledge graphs \citep{markowitz2025kgllmbench}. Two contemporaneous efforts go further and act on the choice: CFPO \citep{liu2025beyond} searches jointly over prompt content and format per task, and DynamicGTR \citep{wei2026dynamicgtr} selects among textual and visual graph topology representations for vision-language models on graph QA. These works establish representation as a first-order factor, but each probes robustness, advocates a single canonical format, or optimizes the choice outside the language-only algorithmic-reasoning setting studied here. GT Bench instead measures representation sensitivity systematically across 44 task--structure settings, four standard representations, and eight models, and GTA turns the choice into an \emph{adaptive}, per-instance selection coupled with plan-decompose-execute, rather than seeking one fixed best format.

Contextual robustness extends beyond graph serialization. Adaptive Distraction uses automated tree search to construct contextual distractions intended to preserve the original question and answer \citep{wang2025adaptivedistraction}, while Cross-Lingual Pitfalls searches for bilingual question pairs that expose inconsistent performance across languages \citep{xu2025crosslingualpitfalls}. Babel Tower examines cross-lingual differences in forward and reverse causal reasoning \citep{wang2025babel}, and Word Form Matters studies semantic reconstruction under scrambled internal letter order \citep{wang2025wordform}. These studies expose sensitivity to how information is presented. GT Bench isolates the effect of encoding a fixed graph and task in different formats.

The use of contextual evidence can also be checked during reasoning. ContextGuard organizes self-auditing around evidence and constraints \citep{jin2026contextguard}, and premise verification decomposes a question into checkable propositions before answering \citep{qin2026premise}. Related benchmarks examine complementary aspects of multi-step reasoning: FineLogic separates answer correctness from intermediate logical soundness \citep{zhou2025finelogic}, while Origami and JigShape test geometric planning and constraint satisfaction \citep{agarwal2026origami,li2026jigshape}. GT Bench supplies explicit structural constraints and algorithmically verifiable targets for its graph tasks.

Benchmark construction and diagnostic resolution further affect what an evaluation reveals. ProbeLLM searches for failure-revealing inputs \citep{huang2026probellm}; DataGen supports controllable synthetic dataset construction \citep{huang2025datagen}; and RefineLab selects data-refinement operations under a token budget \citep{luo2026refinelab}. For reasoning traces, AgentAuditor resolves disagreements through local verification within reasoning trees \citep{yang2026agentauditor}. GT Bench combines parameterized generation with computed reference answers, while reporting the separate plan-alignment diagnostic in \autoref{sec:generator_stability}.

The difficulty of graph tasks for transformers also has a theoretical basis. \citet{sanford2024understanding} place graph algorithms in a representational hierarchy, and \citet{saparov2025transformers} identify graph-search limitations tied to training distributions and graph size. Complementary work on formal verification examines errors in LLM-generated abstract-interpretation steps \citep{mitchell2025formal}. These results motivate explicit evaluation of multi-step reasoning and scaling; the effect of GTA's decomposition is measured by our own ablations (\autoref{tab:ablation_study_results}).

\textbf{What GT Bench adds.}
GT Bench targets multi-step algorithmic reasoning under realistic structural and computational constraints. It spans a broad set of classical graph problems and provides multiple input modalities (natural/structured language, adjacency list/matrix) to stress-test language-based reasoning without relying on code execution or simple lookup. To make the positioning concrete, we summarize key differences with representative benchmarks:

\begin{table}[t]
\centering
\small
\setlength{\tabcolsep}{5pt}
\begin{tabularx}{\textwidth}{@{}>{\raggedright\arraybackslash}p{2.4cm}>{\raggedright\arraybackslash}p{2.55cm}>{\raggedright\arraybackslash}p{2.45cm}>{\raggedright\arraybackslash}X@{}}
\toprule
\textbf{Benchmark} & \textbf{Task coverage} & \textbf{Reported scale} & \textbf{Input and execution protocol} \\
\midrule
GT Bench (Ours) & 24 problems; 44 settings & 105,600 encoded examples & Four equivalent graph encodings; language-only answers; easy/hard subsets. \\
GraCoRe & 19 tasks & 5,140 graphs & Pure and heterogeneous graphs; direct answers; node-order and semantic-feature analyses. \\
NLGraph & 8 tasks & 29,370 problems & Natural-language graphs; direct answers; task-specific difficulty subsets and prompting studies. \\
GraphOmni & 6 core tasks; 2 extensions & 241,726 core queries & Seven serializations and nine prompt schemes; three difficulty levels; adaptive serialization/prompt selection. \\
GraphAlgorithm & 239 problems & 3,041 test instances & Competition problem descriptions; algorithm reasoning followed by generated and executed code. \\
\bottomrule
\end{tabularx}
\caption{Reported scope and evaluation protocols of graph benchmarks: GraCoRe \citep{yuan2024gracore}, NLGraph \citep{wang2023can}, GraphOmni \citep{xu2026graphomni}, and GraphAlgorithm \citep{hu2025rethinking}. Scale units differ across sources and are stated explicitly. NLGraph's count is its extended set; GraphOmni's core count excludes its two additional NP-hard stress tests.}
\label{tab:benchmark_comparison}
\end{table}

As \autoref{tab:benchmark_comparison} shows, GT Bench combines classical problem coverage, equivalent graph encodings, and language-only answers. Its representation study is paired with an agent framework and cross-benchmark transfer evaluation.

\textbf{Relation to GraphOmni and GraphAlgorithm.} GraphOmni \citep{xu2026graphomni} is closely related: both study representation sensitivity and adaptive selection. GraphOmni covers six core tasks with two additional NP-hard stress tests and jointly optimizes serialization and prompting. GT Bench covers 24 classical problems in 44 task--structure settings, and GTA combines representation selection with algorithmic plan--decompose--execute scaffolding. This decomposition adds 7.0 points on GT-E over the Selector-only configuration (\autoref{tab:ablation_study_results}), with further validation on GraCoRe and NLGraph. GraphAlgorithm \citep{hu2025rethinking} instead evaluates algorithm design followed by code generation and execution; its Simple-RTC method reports 92.7--99.9\% accuracy on five existing benchmarks. This protocol measures executable algorithmic solutions, while GT Bench evaluates answers produced in language. AlgoWorlds \citep{xu2026algoworlds} extends tool-mediated evaluation to globally optimal combinatorial decisions in partially observed environments, sharing algorithmically verifiable objectives with GT Bench while using a different observation and execution interface.

\subsection{Attempts to Solve Graph-Theoretic Problems with LLMs}
\label{appendix:atsgtpwl}
Several recent systems attempt to apply LLMs to graph tasks, but most do not directly target language-based algorithmic reasoning over graphs.

\textbf{Graph Agent} \citep{wang2023graph} uses explicit inductive and deductive reasoning for node classification and link prediction. LLM-GOOD \citep{xu2025llmgood} combines LLM-assisted annotation with a lightweight graph neural network for few-shot out-of-distribution detection on text-attributed graphs. These prediction tasks differ from executing classical graph algorithms in language.

Relational representation learning distinguishes constructing a useful graph representation from executing an algorithm on a given graph. NQE embeds multi-hop logical queries on n-ary knowledge graphs \citep{luo2023nqe}; DHGE couples instance-level and ontology-level representations \citep{luo2023dhge}; and HAHE combines global and local attention for hyper-relational knowledge graphs \citep{luo2023hahe}. Text2NKG constructs n-ary knowledge graphs from text \citep{luo2024text2nkg}. GT Bench starts from an explicitly specified graph and evaluates the requested computation across equivalent encodings.

Recent work on text-attributed graphs connects language models to data selection and distribution shifts. GOE-LLM generates outlier exposure data for graph out-of-distribution detection \citep{xu2025goellm}, and LEGO-Learn studies label-efficient learning on open-set graphs \citep{xu2025legolearn}. TAG-AD evaluates contextual and structural anomalies and introduces retrieval-assisted zero-shot detection \citep{xu2025tagad}. Together with LLM-GOOD, these prediction-oriented methods separate the contributions of semantic content, graph structure, and model choice, complementing the algorithmic setting of GT Bench.

Graph-learning robustness also distinguishes prediction stability from explanation stability. Small structural perturbations can change graph explanations even when predictions remain stable \citep{li2024gnnfragile}, motivating certification methods such as XGNNCert \citep{li2025xgnncert}. These studies perturb the graph itself. GT Bench keeps the graph invariant, so differences across its encodings arise without changing the underlying problem.

Explicit intermediate structure is also relevant to reasoning fidelity. CDCR-SFT trains models to construct causal directed acyclic graphs and reason over them \citep{li2026cdcr}. Work on cooperative rationalization shows that an accurate predictor can still exploit spurious correlations introduced by rationale selection, including on graph-classification data \citep{liu2025advrat}. In classical program analysis, incremental algebraic program analysis reuses and updates symbolic path summaries after program changes \citep{zhou2025incrementalapa}. These studies connect structured intermediate states to causal reasoning, explanation fidelity, and path computation, respectively. They motivate evaluating the stages of a reasoning pipeline separately rather than inferring their reliability from final accuracy alone.

\textbf{GraphAgent-Reasoner} \citep{hu2024scalableaccurategraphreasoning} distributes node-centric computation across an agent network, with a master LLM coordinating the algorithm and summarizing final states. Its distributed execution protocol differs from GTA's single frozen executor.

\textbf{GraphTeam} \citep{li2025graphteamfacilitatinglargelanguage} orchestrates multiple agents in a workflow (Original Question $\to$ Question Agent $\to$ Search Agent $\to$ Coding Agent), with a fallback Reasoning Agent when code fails. This \emph{code-first} pipeline is effective for analysis workflows, but it \emph{bypasses} intrinsic language-based graph reasoning by delegating problem solving to generated code and execution. Our objective is orthogonal: evaluate and enhance an LLM’s \emph{intrinsic} algorithmic reasoning over graph structure without relying on external code execution.

Beyond these, training- or alignment-centric approaches such as \textbf{GUNDAM} \citep{ouyang2024gundamaligninglargelanguage}, \textbf{GraphThought} \citep{huang2025graphthoughtgraphcombinatorialoptimization}, \textbf{GraphWiz} \citep{chen2024graphwizinstructionfollowinglanguagemodel}, and \textbf{GCoder} \citep{zhang2024gcoderimprovinglargelanguage} explore instruction tuning, preference alignment, or thought-generation for graph tasks, and \textbf{GraphInstruct} \citep{luo2024graphinstruct} contributes an instruction corpus with intermediate reasoning steps across 21 graph tasks. This line remains active in the reasoning-model era: \textbf{G1} \citep{guo2025g1} applies reinforcement learning on a synthetic graph corpus so that a small model surpasses far larger ones, \textbf{NPG-Muse} \citep{wang2025npgmuse} elicits long chain-of-thought reasoning by training on NP-hard graph problems, \citet{zhang2025improving} continue pretraining on a 10.9B-token graph-problem corpus and report transfer to mathematical and logical reasoning, \citet{zhang2025generalizable} compare solution- and process-based post-training rewards on synthetic graph problems and evaluate transfer to real-world tasks, \textbf{GraphTool-Instruction} \citep{wang2024graphtool} decomposes reasoning into subtasks that call external graph tools, and further work fine-tunes LLMs as end-to-end combinatorial-optimization solvers \citep{jiang2025e2eco} or as reasoning-then-coding agents \citep{hu2025rethinking}. These are promising directions, yet they typically modify or fine-tune the \emph{executor} model itself or rely on tool/code execution. In contrast, our GTA explicitly (i) keeps the executor \emph{frozen} and (ii) compels the model to \emph{solve by reasoning} through an adaptive representation selector and a lightweight plan–decompose–execute pipeline. This separation isolates the benefit of better \emph{reasoning and representation choices}, rather than changes in model capacity.

A further alternative changes the graph interface. GraphLLM \citep{chai2023graphllmboostinggraphreasoning} learns graph-enhanced prefixes for a frozen LLM, GraphToken \citep{perozzi2024let} learns continuous soft-prompt encodings, and G-Retriever \citep{he2024gretriever} couples a graph encoder with retrieval for textual graph question answering. These methods require embedding-space access or an auxiliary encoder. ReMindRAG \citep{hu2025remindrag} instead guides knowledge-graph traversal with an LLM and reuses traversal experience for efficient retrieval. GTA selects among textual encodings of the complete problem graph and supplies algorithmic decomposition to a frozen executor.

\subsection{LLM-Based Agent Frameworks}
The view of LLMs as agents that plan, decompose, and act has gained momentum. Chain-of-Thought prompting \citep{wei2022chain} exposes intermediate reasoning steps, ReAct \citep{yao2023react} interleaves reasoning with actions, and Toolformer \citep{schick2023toolformer} learns when to invoke tools. For graphs, Graph-CoT \citep{jin2024graph} alternates reasoning, graph interaction, and graph execution. AFlow \citep{zhang2024aflow} searches over workflows of LLM calls, whereas DyFlow \citep{wang2025dyflowdynamicworkflowframework} uses a designer--executor architecture to revise subgoal plans in response to intermediate execution feedback.

Adaptive controllers can also alter how reasoning is carried out. AdaReasoner learns task-dependent reasoning configurations through an auxiliary policy \citep{wang2025adareasoner}; InfoQA combines capacity-aware decomposition with pruning of earlier reasoning traces for multi-hop question answering \citep{wan2026infoqa}; and Learning to Deliberate learns policies over persist, refine, and concede actions in multi-agent reasoning \citep{yang2026deliberate}. CoAct studies parallel agent execution through contrastive task allocation \citep{peng2026coact}. GTA specializes adaptation to graph encoding and algorithmic decomposition. A distinct agents-as-topology setting is studied by AgentsNet \citep{grotschla2025agentsnet}, where node-level LLM agents coordinate on distributed graph problems and performance degrades as the communication network grows.

Workflow adaptation can be learned at several levels. SWIFT transfers structural priors and workflow demonstrations across tasks to synthesize workflows with a single generation pass \citep{du2026swift}. OASES jointly trains search and state-evaluation policies to align intermediate search feedback with final outcomes \citep{zhang2026oases}. Unity-MAS converts workflows into role-conditioned trajectories for multi-agent reinforcement learning \citep{chen2026unitymas}, while MAESTRO trains execution and synthesis agents with conditional policy optimization \citep{yang2025maestro}. Self-Compressing MAS learns to reduce redundant reasoning content \citep{chen2026selfcompress}. These approaches optimize workflow structure, credit assignment, or communication. GTA uses a fixed high-level workflow and learns the representation choice and decomposition for its graph tasks.

Decomposition makes intermediate interfaces explicit across several agent settings. CogWriter combines hierarchical planning, parallel drafting, and review for long-form writing \citep{wan2025cogwriter}; AD-AGENT organizes data analysis, model selection, and code execution for anomaly detection \citep{yang2025adagent}; and QUITE uses a finite-state multi-agent workflow for SQL rewriting with semantic and execution checks \citep{song2026quite}. These examples connect task decomposition to the checks available at each stage, whereas GTA executes the resulting graph-reasoning steps in language.

Control policies further determine how plans respond to observations. ASTRO learns planning strategies for non-cooperative dialogue \citep{hu2025astro}. Three-Step Nav combines global planning, local subgoals, and retrospective checking \citep{zheng2026threestep}; Ponder separates GUI-action interpretation from action localization \citep{wang2025ponder}; and ManipLVM-R1 uses affordance information and verifiable trajectory rewards for manipulation \citep{song2026maniplvm}. These systems illustrate the separation of high-level planning, intermediate representations, and execution in interactive tasks.

Reusable skills and memory offer another form of adaptation. Recipes for Agents describes skills as reusable procedural knowledge and examines their construction, composition, and evaluation \citep{xing2026recipes}. HyperSkill represents skills and trajectories in a hypergraph for retrieval and reuse \citep{xu2026hyperskill}; SkillGen derives reusable skills from contrasting successful and unsuccessful trajectories \citep{ma2026skillgen}. RaMem preserves the contextual conditions needed to interpret retrieved memories as evidence \citep{yang2026ramem}. MemoHarness learns an external controller around a frozen model \citep{huang2026memoharness}, and MetaCollab combines autonomous execution, selective human intervention, and continual improvement \citep{yang2026metacollab}. These systems distinguish persistent experience from the current task state. GTA's cross-benchmark evaluation instead tests transfer of its learned auxiliary components without benchmark-specific retraining.

Selection matters even when the available components are fixed. MetaTool evaluates whether an LLM should use a tool and which one it should choose \citep{huang2024metatool}, while AD-LLM evaluates model selection in anomaly detection \citep{yang2025adllm}. MetaOOD and M3OOD use historical performance and task descriptors to select out-of-distribution detectors \citep{qin2025metaood,qin2026m3ood}; FlexRouter studies complementary model sets for flexible LLM routing \citep{wei2026flexrouter}. GTA fixes the executor and learns which graph encoding is most effective for the current instance.

Auxiliary components can also differ in the supervision used to train them. CoAct combines self-rewarding with active preference annotation \citep{xu2026coactpref}; it is distinct from the contrastive task-allocation framework of the same name \citep{peng2026coact}. TerminalTraj generates verifiable terminal-agent trajectories \citep{wu2026terminaltraj}, and Context-CoT constructs context-grounded reasoning traces for downstream training \citep{jin2026contextcot}. RMO reshapes reward-margin distributions \citep{ru2026rmo}, while DOG-DPO selects preference data through geometric coverage \citep{nian2026dogdpo}. These methods address the source or quality of supervision, which is relevant to GTA's preference-trained selector and decomposer.

Execution structure provides a complementary basis for diagnosis. GRADE represents dependencies and execution structure in agent runs \citep{zhao2026grade}, and RiskLab varies interaction topology, protocols, agents, and tasks to study multi-agent failures \citep{jiang2026risklab}. GTA evaluates the effects of representation selection and decomposition through component ablations within its specified workflow.

Our Graph Theory Agent specializes this broader agent-design space to algorithmic graph reasoning. It combines an adaptive graph representation selector with plan--decompose--execute scaffolding around a frozen executor. The resulting design makes representation choice and decomposition directly testable through component ablations and transfer to graph benchmarks with different task distributions.

\section{GT Bench: Graph Theory Benchmark}
\label{sec:graphbench_dataset}

To rigorously evaluate LLM algorithmic reasoning on graphs, we constructed GT Bench to probe multi-step inference across diverse graph structures and graph-theoretic problems. Detailed task definitions and ground-truth algorithms are provided in \autoref{appendix:task_definitions_algorithms}.

\subsection{Graph Characteristics}
The graphs within GT Bench are primarily categorized by edge density into sparse graphs, where the number of edges \(E\) scales linearly with the number of vertices \(V\) (i.e., \(E = O(V)\)), and dense graphs, where \(E\) approaches the quadratic maximum (\(E = O(V^2)\)). A third category comprises trees, defined as connected graphs satisfying \(E = V-1\). Depending on specific task requirements, graphs are generated with varying properties, such as weighted or unweighted edges, and may be connected or disconnected. Specific topological structures like star graphs are also implicitly generated within these categories.

\begin{figure*}[t]
    \centering
    \includegraphics[width=\linewidth]{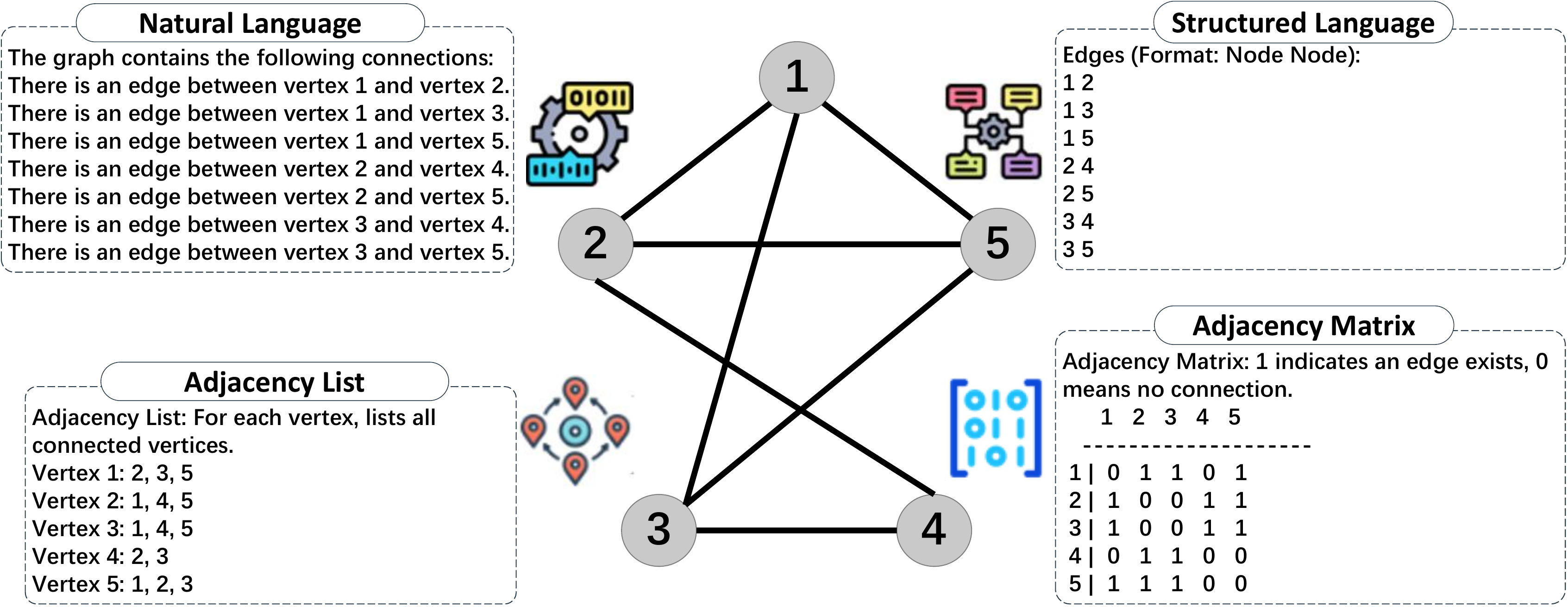} 
    \caption{Example of an undirected graph represented using the four input modalities evaluated in this work: Natural Language, Structured Language, Adjacency List, and Adjacency Matrix.}
    \label{fig:input_formats}
    \vspace{-2em}
\end{figure*}

\subsection{Input Representations}
A distinctive feature of GT Bench is that every problem instance is provided in four input modalities, enabling a controlled study of how representation choice affects reasoning performance. The four modalities are: \textbf{Natural Language (NL)}, a descriptive prose format detailing vertices, edges, and weights where applicable; \textbf{Structured Language (SL)}, which employs a more templated approach using keywords and indentation to articulate graph structure; the \textbf{Adjacency Matrix (AM)}, a matrix-based representation where entries denote edge existence or their associated weights; and the \textbf{Adjacency List (AL)}, which lists the neighbors and corresponding weights for each vertex. \autoref{fig:input_formats} offers a visual comparison of these four representations applied to an exemplary small graph.

\subsection{Task Coverage and Dataset Composition}
GT Bench encompasses 24 classical graph problems, ranging from connectivity checks to optimization problems such as MCMF. Twenty problem types are evaluated on both sparse and dense graphs, and four are tree-specific, yielding 44 task--structure settings. Graph sizes are governed by a prompt-fit budget: node ranges are calibrated per task family so that every instance, under any of the four representations, fits comfortably within the context windows of contemporary LLMs, with tighter ranges for dense families whose edge count grows quadratically. The graphs remain large enough that answers require genuine multi-step reasoning rather than lookup.
\autoref{tab:graphbench_tasks} details the resulting node counts and instance numbers for each problem type.
Collectively, the dataset comprises 105{,}600 examples across the four representations, which are further stratified into \textit{easy} (GT-E) and \textit{hard} (GT-H) subsets to accommodate varying levels of difficulty, where details can be found in \autoref{appendix:task_definitions_algorithms}.

The generation of the entire dataset, including graph structures, problem formulations, and solution verifications, is fully automated through algorithmic procedures.
This generation algorithm also supports adjusting graph scales to produce new data instances; the evaluated size ranges and scope are detailed in \autoref{app:evaluation_scope}. Ground truth solutions are systematically derived using canonical algorithms specific to each task, thereby ensuring correctness and eliminating the need for manual annotation or labeling. The accompanying repository provides sample instances and the configurable generation pipeline.

\begingroup
\definecolor{headerbg}{RGB}{27,71,109}
\definecolor{rowbg}{RGB}{240,247,255}
\definecolor{rowbg2}{RGB}{225,235,245}
\definecolor{headerfg}{RGB}{255,255,255}

\begin{table}[p]
\centering
\caption{Overview of the 24 problem types in GT Bench. Example counts include the four input representations; sparse/dense variants and tree tasks together form 44 task--structure settings.}
\label{tab:graphbench_tasks}
\rowcolors{2}{rowbg}{rowbg2}
\renewcommand{\arraystretch}{1.15}
\resizebox{\textwidth}{!}{%
\begin{tabular}{>{\raggedright\arraybackslash}p{4cm} >{\raggedright\arraybackslash}p{8.5cm} >{\centering\arraybackslash}p{2cm} >{\centering\arraybackslash}p{2cm}}
\rowcolor{headerbg}
\textcolor{headerfg}{\textbf{Task}} 
    & \textcolor{headerfg}{\textbf{Description}} 
    & \textcolor{headerfg}{\textbf{Nodes}} 
    & \textcolor{headerfg}{\textbf{Examples}} \\
\toprule
Connectivity & Determine whether two vertices are connected & 15--40 & 4800 \\
Bipartiteness Check & Determine whether the graph is bipartite & 15--40 & 4800 \\
Minimum Cycle Length & Find the length of the smallest cycle in the graph & 15--40 & 4800 \\
Maximum Clique Size & Compute the size of the largest clique & 15--40 & 4800 \\
Maximum Independent Set & Compute the size of the largest independent set & 15--40 & 4800 \\
Eulerian Path & Determine whether the graph contains an Eulerian path & 15--40 & 4800 \\
Eulerian Circuit & Determine whether the graph contains an Eulerian circuit & 15--40 & 4800 \\
Hamiltonian Path & Determine whether the graph contains a Hamiltonian path & 15--40 & 4800 \\
Hamiltonian Circuit & Determine whether the graph contains a Hamiltonian circuit & 15--40 & 4800 \\
Biconnected Components & Count the number of biconnected components & 11--20 & 4800 \\
Bridge Count & Count the number of bridges (cut edges) & 11--20 & 4800 \\
Triangle Count & Count the number of triangles (3-node cycles) & 11--13 & 4800 \\
Cycle Count & Count the total number of cycles in the graph & 6--10 & 4800 \\
Spanning Tree Count & Count the number of possible spanning trees & 6--10 & 4800 \\
Shortest Path Length & Compute the shortest path between two nodes & 14--20 & 4800 \\
Minimum Spanning Tree & Compute the total weight of the minimum spanning tree & 14--20 & 4800 \\
Second Minimum Spanning Tree & Compute the weight of the second-best minimum spanning tree & 11--20 & 4800 \\
Tree Diameter & Compute the diameter (longest shortest path) of the tree & 29--31 & 2400 \\
Tree Centroid & Find the centroid node with the smallest index & 29--60 & 2400 \\
Lowest Common Ancestor & Find the lowest common ancestor (root at node 1) & 29--60 & 2400 \\
Tree Max Independent Set & Compute the size of the maximum independent set in a tree & 29--60 & 2400 \\
Maximum Flow & Compute the maximum flow from source to sink & 11--20 & 4800 \\
Minimum Cut & Compute the capacity of the minimum cut (source to sink) & 11--20 & 4800 \\
Min-Cost Max-Flow & Compute the minimum-cost maximum flow & 9--15 & 4800 \\
\bottomrule
\end{tabular}
}
\end{table}
\vspace{-0em}
\endgroup

\section{Benchmarking LLMs on GT Bench}
\label{sec:bench_findings}

We evaluate eight LLMs on GT Bench: three proprietary models, \texttt{GPT-4o-mini} \citep{openai2024gpt4omini}, \texttt{GPT-4o} \citep{hurst2024gpt}, and \texttt{o3-mini} \citep{o3mini2025}, and five open-source models, \texttt{Llama-3.1-8B} \citep{meta2024llama31_8b}, \texttt{Llama-3.3-70B} \citep{meta2024llama31_70b}, \texttt{Phi-4} \citep{phi4_2024}, \texttt{QwQ-32B} \citep{qwq32b}, and \texttt{DeepSeek-R1} \citep{guo2025deepseek}. Each model is evaluated on 10{,}000 instances across the four representations, using one inference run per configuration at temperature 0.01. Model versions and settings appear in \autoref{appendix:experimental_details}, with evaluation scope and statistical interpretation in \autoref{app:evaluation_scope}.

\subsection{Representation Sensitivity}
\label{sec:model_performance_gtbench}
\label{subsec:representation_sensitivity}

\begin{table}[ht]
\vspace{-1em}
\centering
\begingroup
\definecolor{headerbg2}{RGB}{230,250,236}
\definecolor{rowone2}{RGB}{245,252,247}
\definecolor{rowtwo2}{RGB}{240,246,242}
\setlength{\arrayrulewidth}{0.5pt}
\arrayrulecolor{gray!50}
\rowcolors{2}{rowone2}{rowtwo2}
\setlength{\tabcolsep}{5pt}

\caption{Accuracy (\%) on GT Bench across selected tasks and input representations, averaged across models. Average and Range summarize the four representations and are computed before rounding.}
\label{tab:task_representation_accuracy}
\begin{tabularx}{\textwidth}{>{\columncolor{headerbg2}}l >{\columncolor{headerbg2}}l *{6}{>{\centering\arraybackslash}X}}
\toprule
\rowcolor{headerbg2}
\textbf{Task} & \textbf{Graph Type} & \textbf{NL} & \textbf{SL} & \textbf{AM} & \textbf{AL} & \textbf{Average} & \textbf{Range} \\
\midrule
Bipartite Check        & Sparse & \textbf{89.9} & 83.6 & 75.9 & 77.0 & 81.6 & 14.0 \\
Tree Centroid          & Tree   & 56.3 & \textbf{60.8} & 42.6 & 51.1 & 52.7 & 18.3 \\
Maximum Clique         & Dense  & 24.7 & 22.1 & \textbf{38.6} & 34.9 & 30.1 & 16.5 \\
Hamiltonian Circuit    & Dense  & 81.4 & 81.0 & 83.0 & \textbf{88.3} & 83.4 & 7.3 \\
\bottomrule
\end{tabularx}
\endgroup
\vspace{-0em}
\end{table}

\textbf{Different models exhibit distinct preferences for input representations.}
As shown in \autoref{tab:model_representation_accuracy}, no single input format is universally optimal. For example, \texttt{llama-3.1-8B} favors AM (29.4\%), while \texttt{phi-4} performs best with SL (44.9\%). Larger models like \texttt{gpt-4o} achieve peak accuracy with AL (47.6\%), whereas \texttt{deepseek-r1} prefers AM (82.5\%). The preference is model-specific, so any evaluation or deployment that fixes one format across models will flatter some and shortchange others.

\needspace{16\baselineskip}
\begin{wraptable}{r}{0.5\textwidth}
\vspace{-1.4em}
\begingroup
\definecolor{headerbg}{RGB}{230,245,255} 
\definecolor{rowone}{RGB}{245,251,255}   
\definecolor{rowtwo}{RGB}{240,247,255}   
\setlength{\arrayrulewidth}{0.5pt}
\arrayrulecolor{gray!50}
\rowcolors{2}{rowone}{rowtwo} 
\centering
\caption{Accuracy (\%) of different models on GT Bench under four input representations.}
\label{tab:model_representation_accuracy}
\begin{tabular}{>{\columncolor{headerbg}}lcccc}
\toprule
\rowcolor{headerbg}
\textbf{Model} & \textbf{NL} & \textbf{SL} & \textbf{AM} & \textbf{AL} \\
\midrule
Llama-3.1-8B      & 29.0 & 27.4 & \textbf{29.4} & 28.5 \\
Phi-4             & 43.7 & \textbf{44.9} & 38.8 & 42.7 \\
GPT-4o-mini       & 36.2 & 33.4 & 37.8 & \textbf{40.7} \\
Llama-3.3-70B     & \textbf{44.9} & 41.1 & 40.7 & 44.1 \\
GPT-4o            & 45.5 & 43.7 & 45.3 & \textbf{47.6} \\
QwQ-32B           & 63.1 & 63.1 & 62.1 & \textbf{71.3} \\
DeepSeek-R1       & 80.9 & 80.6 & \textbf{82.5} & 80.9 \\
o3-mini           & 87.4 & 89.9 & 89.0 & \textbf{90.2} \\
\bottomrule
\end{tabular}
\endgroup
\vspace{-1em}
\end{wraptable}

\textbf{The optimal input representation is highly dependent on the specific task and graph structure.}
\autoref{tab:task_representation_accuracy} illustrates this dependency with selected tasks from GT Bench. For instance, NL representation (89.9\%) is most effective for Bipartite Check on sparse graphs, whereas AM (38.6\%) is superior for Maximum Clique on dense graphs. Tree-centric tasks like Tree Centroid benefit from SL (60.8\%), and path-oriented problems such as Hamiltonian Circuit on dense graphs achieve highest accuracy with AL (88.3\%). Task-level results for these examples appear in \autoref{appendix:experimental_results} and form the empirical basis of GTA's heuristic selector.

Which representation wins is not arbitrary. \autoref{tab:best_formats} in the appendix tabulates the empirically best format for every task--graph-type combination, and four regularities emerge:

\textbf{NL} suits tasks on sparse graphs, where little structural parsing is required and the descriptive prose plays to the model's linguistic strengths.

\textbf{SL} dominates on trees; its templated, indented format mirrors the hierarchy that tree algorithms traverse.

\textbf{AM} wins on dense graphs, where the matrix lays out every potential pairwise connection at once, making global properties easier to read off.

\textbf{AL} is strongest for pathfinding and neighbor-iteration tasks at any density, since it lists exactly the edges an algorithm would walk.

Representation sensitivity persists in the strongest evaluated reasoning models, as shown in \autoref{tab:model_representation_accuracy}. The gap between a model's best and worst format is 2.0 points for \texttt{Llama-3.1-8B}, which sits near floor under every format, widens to 6.1 for \texttt{Phi-4}, 7.3 for \texttt{GPT-4o-mini}, and 9.2 for \texttt{QwQ-32B}, and only then narrows to 2.8 for \texttt{o3-mini} and 1.9 for \texttt{DeepSeek-R1}. Sensitivity is thus largest in the capable-but-unsaturated regime where benchmark performance is most informative, and it shrinks, but does not vanish, for the strongest models. \autoref{sec:case_studies} complements these aggregates with paired case studies in which the same model, given the identical instance, fails under one representation and succeeds under another.

\subsection{Performance across Difficulty Levels}
\label{sec:appendix_gtbench_results}

Aggregate accuracy hides how sharply GT Bench separates models once difficulty is taken into account. Instances are stratified as follows: four intrinsically demanding problem types (Min-Cost Max-Flow, Cycle Count, Spanning Tree Count, and Biconnected Components) are always Hard, two elementary ones (Connectivity and Bipartiteness) are always Easy, and for every other task, instances on sparse graphs or trees count as Easy while instances on dense graphs count as Hard. \autoref{tab:appendix_model_easy_hard_accuracy} reports per-model accuracy under this split.

\begin{table}[ht]
\centering
\caption{Accuracy (\%) of different LLMs on the Easy and Hard subsets of GT Bench. Easy and Hard subsets are defined based on task type and graph structure as detailed in \autoref{sec:appendix_gtbench_results}.}
\label{tab:appendix_model_easy_hard_accuracy}
\begin{tabular*}{\textwidth}{@{\extracolsep{\fill}}lcc@{}}
\toprule
\textbf{Model} & \textbf{Easy Task Accuracy (\%)} & \textbf{Hard Task Accuracy (\%)} \\
\midrule
\texttt{Llama-3.1-8B}  & 38.13 & 19.05 \\
\texttt{GPT-4o-mini}   & 44.15 & 27.47 \\
\texttt{Phi-4}         & 53.51 & 32.97 \\
\texttt{Llama-3.3-70B} & 58.19 & 30.40 \\
\texttt{GPT-4o}        & 60.54 & 28.94 \\
\texttt{QwQ-32B}           & 85.62 & 38.46 \\
\texttt{DeepSeek-r1}   & 96.32 & 64.10 \\
\texttt{o3-mini}       & 98.66 & 75.09 \\
\bottomrule
\end{tabular*}
\end{table}

Two aspects of the split stand out. All eight models drop sharply from Easy to Hard; \texttt{QwQ-32B}, for example, falls by 47 points, from 85.62\% to 38.46\%. The model ordering also changes: \texttt{GPT-4o} edges out \texttt{Llama-3.3-70B} on Easy instances (60.54\% versus 58.19\%) yet trails it on Hard ones (28.94\% versus 30.40\%), so rankings drawn from easy tasks alone can mislead. Even the strongest reasoning models remain well below ceiling on the Hard split, and this is the headroom that the methods evaluated in \autoref{sec:experiments} target.
\needspace{14\baselineskip}
\subsection{Does Providing Every Representation Help?}
\label{sec:appendix_all_reprs}

\begin{wraptable}{r}{0.5\textwidth}
\vspace{-1.4em}
\centering
\begingroup
\definecolor{headerbgall}{RGB}{235,245,255}
\definecolor{rowoneall}{RGB}{245,250,255}
\definecolor{rowtwoall}{RGB}{240,246,252}
\setlength{\arrayrulewidth}{0.5pt}
\arrayrulecolor{gray!50}
\rowcolors{2}{rowoneall}{rowtwoall}
\small
\caption{Accuracy (\%) under individual representations vs.\ providing all four representations simultaneously. Best single-representation result for each model is \textbf{bolded}.}
\label{tab:all_reprs}
\vspace{0.3em}
\setlength{\tabcolsep}{4pt}
\begin{tabular}{>{\columncolor{headerbgall}}lcccc|c}
\toprule
\rowcolor{headerbgall}
\textbf{Model} & \textbf{NL} & \textbf{SL} & \textbf{AM} & \textbf{AL} & \textbf{All} \\
\midrule
Llama-3.1-8B & 29.0 & 27.4 & \textbf{29.4} & 28.5 & 22.4 \\
Phi-4        & 43.7 & \textbf{44.9} & 38.8 & 42.7 & 34.6 \\
GPT-4o       & 45.5 & 43.7 & 45.3 & \textbf{47.6} & 40.3 \\
\bottomrule
\end{tabular}
\endgroup
\vspace{-1em}
\end{wraptable}
To test whether providing all four representations simultaneously improves performance, we concatenated all formats into a single prompt. As \autoref{tab:all_reprs} shows, this consistently degrades accuracy relative to the best single format for each model; \texttt{GPT-4o}, for instance, drops from 47.6\% to 40.3\%. Two effects compound. Concatenation inflates the prompt roughly three to four times, which risks context-window overflow on larger graphs, and the structural diversity of four simultaneous formats introduces noise that the model must reconcile rather than exploit. Adaptive selection of a single, well-matched representation is therefore preferable to naively supplying every available encoding.

\subsection{Relationship to General Capabilities}
\label{sec:orthogonality}

To understand whether graph-theoretic reasoning tests capabilities distinct from those measured by existing benchmarks, we compare model performance on GT Bench (Hard subset) with their Elo ratings on Text Arena\footnote{\url{https://lmarena.ai/leaderboard/text}} across four capability dimensions: Coding, MATH, Instruction Following, and Longer Query (\autoref{tab:orthogonality}). Three patterns emerge. First, graph reasoning diverges sharply from general language capabilities. The most telling example is \texttt{QwQ-32B}: despite having the \emph{lowest} scores in Instruction Following (1149) and Longer Query (1176), even lower than the 8B-parameter \texttt{Llama-3.1-8B}, it ranks 3rd on GT-H, substantially outperforming the much larger \texttt{GPT-4o} and \texttt{Llama-3.3-70B}. Second, the top performers on GT-H (\texttt{o3-mini}, \texttt{DeepSeek-R1}, \texttt{QwQ-32B}) are consistently the leaders in Coding and MATH, suggesting that graph-theoretic reasoning is rooted in precise, multi-step algorithmic execution, a capability shared with formal reasoning but distinct from natural language understanding. Third, model scale is not the primary driver of success: smaller, reasoning-optimized models consistently outperform larger generalist models. GT Bench thus probes an axis of capability that general leaderboards do not capture. That this axis is both distinctive and transferable is corroborated by \citet{zhang2025improving}, who find that continued pretraining on graph problems improves mathematical and logical reasoning downstream.

\begin{table}[ht]
\centering
\begingroup
\definecolor{headerbgorth}{RGB}{240,235,250}
\definecolor{rowoneorth}{RGB}{248,246,253}
\definecolor{rowtwoorth}{RGB}{242,238,250}
\setlength{\arrayrulewidth}{0.5pt}
\arrayrulecolor{gray!50}
\rowcolors{2}{rowoneorth}{rowtwoorth}
\caption{GT Bench (Hard) accuracy vs.\ Text Arena Elo ratings across capability dimensions. Models sorted by GT-H performance.}
\label{tab:orthogonality}
\vspace{0.5em}
\setlength{\tabcolsep}{4.5pt}
\small
\begin{tabularx}{\textwidth}{>{\columncolor{headerbgorth}}l *{5}{>{\centering\arraybackslash}X}}
\toprule
\rowcolor{headerbgorth}
\textbf{Model} & \textbf{GT-H} & \textbf{Coding} & \textbf{MATH} & \textbf{Instr. Follow} & \textbf{Longer Query} \\
\midrule
Llama-3.1-8B  & 19.1\% & 1258 & 1194 & 1187 & 1220 \\
GPT-4o-mini   & 27.5\% & 1347 & 1279 & 1288 & 1322 \\
GPT-4o        & 28.9\% & 1367 & 1308 & 1319 & 1328 \\
Llama-3.3-70B & 30.4\% & 1343 & 1299 & 1287 & 1309 \\
Phi-4         & 33.0\% & 1304 & 1268 & 1239 & 1263 \\
QwQ-32B       & 38.5\% & 1385 & 1366 & 1149 & 1176 \\
DeepSeek-R1   & 64.1\% & 1443 & 1398 & 1390 & 1396 \\
o3-mini       & 75.1\% & 1413 & 1387 & 1339 & 1358 \\
\bottomrule
\end{tabularx}
\endgroup
\end{table}

Across these analyses a consistent picture emerges: accuracy swings with the input representation by up to nine points per model and far more on individual tasks; the winning format depends jointly on task, graph structure, and model; supplying every format at once backfires; and the Hard split remains far from saturated even for strong reasoning models. A remedy should therefore choose a suitable representation per instance rather than fix one globally, should support the executor through the multi-step algorithmic work on which unaided models fail, and should do both without retraining the executor, so that improvements reflect better problem presentation rather than added capacity. The agent introduced next is built around exactly these requirements.

\section{Graph Theory Agent Framework}
\label{sec:gta_agent}

The \textbf{Graph Theory Agent (GTA)} turns the diagnosis of \autoref{sec:bench_findings} into a design. It combines two mechanisms: adaptive selection of the input representation for each instance, targeting the sensitivity documented in \autoref{sec:model_performance_gtbench}, and systematic decomposition of the solution process, targeting the multi-step failures behind the Easy--Hard gap. Crucially, GTA operates under a \emph{frozen executor} paradigm: the base LLM is never trained or fine-tuned on GT Bench data. Only lightweight auxiliary modules (the Selector and Decomposer) are optimized, ensuring that performance gains are attributable to the framework's reasoning strategy rather than changes in the executor's intrinsic capabilities. This distinguishes GTA from fine-tuning approaches such as GraphWiz, which modify the executor itself.
GTA tackles a graph problem instance $P = (G, T)$ (where $G$ is the graph and $T$ the task) via a structured pipeline:
\begin{equation}
    P \xrightarrow{\text{Selector}} P' \xrightarrow{\text{Generator}} \mathcal{A} \xrightarrow{\text{Decomposer}} \{s_1, ..., s_k\} \xrightarrow{\text{Executor}} \text{Solution}
\end{equation}
The pipeline begins with the \textbf{Input Representation Selector} choosing an optimal format $R^*$ for graph $G$ and task $T$, yielding $P'=(G_{R^*}, T)$. The \textbf{Algorithm Generator} then formulates a high-level strategy $\mathcal{A}$, which the \textbf{Algorithm Decomposer} refines into a sequence of manageable sub-steps $\{s_1, ..., s_k\}$. Finally, an \textbf{Executor} LLM works through these sub-steps to compute the solution.

\subsection{Input Representation Selector}
\label{subsec:gta_selector}

The Input Representation Selector is the first component in the GTA pipeline. Its role is to choose the most effective format from the available options $\mathcal{R} = \{ \text{NL, SL, AM, AL} \}$ to present the input graph $G$ for a given task $T$. The objective is to select the representation $R^*$ that maximizes the likelihood of successful downstream reasoning by the Executor LLM. This selection can be formalized as $\mathcal{S}: (G, T) \mapsto R^*$, where $R^* \in \mathcal{R}$ is determined based on features of $G$ (e.g., size, density) and the requirements of $T$. We explore two implementations for this adaptive selector:

\textbf{1. Heuristic-Based Selector:} This approach employs a rule-based function derived from our empirical findings (\autoref{tab:best_formats}). It maps observable graph and task characteristics to the representation that demonstrated the strongest average performance. Its primary advantage is computational efficiency.

\textbf{2. Learned Selector:} Alternatively, this selector is a dedicated language model, fine-tuned for representation selection. This allows optimization for a \textit{specific} downstream Executor LLM, trained to predict the representation most likely to yield successful task completion by that target model using historical data of problem instances, representations, and outcomes. The learned selector's advantage over the heuristic is most pronounced in two scenarios: (i)~when the executor LLM deviates from general statistical trends (for example, reasoning-specialized models like \texttt{QwQ-32B} may exhibit different parsing preferences than generalist models), and (ii)~for graphs in ambiguous structural regimes (e.g., medium density with complex community structures), where the optimal representation depends on finer-grained correlations between task requirements and the executor's specific capabilities. Henceforth, unless otherwise specified, references to GTA will imply the use of the Learned Selector.

By tailoring the input format per instance, the Selector confronts the representation sensitivity documented in \autoref{sec:bench_findings} and hands the downstream components a graph encoding the executor is well placed to reason over.

The Learned Selector is also the component that carries GTA's generalization, along three axes. Along the \emph{model} axis, a single Selector trained on the Phi-4 pipeline transfers to unseen Executors without retraining (\autoref{tab:gta_executor_comparison}). Along the \emph{task} axis, it extends to new problem families by fine-tuning on a small set of (task, graph, representation, outcome) preference pairs through the same procedure of \autoref{subsec:gta_training_workflow}. Along the \emph{representation} axis, the candidate set $\mathcal{R}$ can be enlarged with additional formats without any architectural change, since the Selector consumes only $(G,T)$ features and returns a choice from $\mathcal{R}$. The static Heuristic-Based Selector, by contrast, is a lookup derived from GT Bench statistics and is best viewed as an economical fallback that would need re-derivation outside its coverage.

\subsection{Algorithm Generator}
\label{subsec:gta_generator}

Following the selection of an optimal representation $R^*$, the \textbf{Algorithm Generator} formulates a \textit{high-level} strategic plan $\mathcal{A}$. This component, an off-the-shelf LLM, is prompted with the task description $T$ and the graph $G_{R^*}$ in its chosen format. Its role is conceptual: to establish the overarching algorithmic approach and outline key reasoning phases. It explicitly avoids generating detailed instructions or executable code. The resultant plan $\mathcal{A}$ serves as a strategic blueprint, ensuring a logically sound methodology and providing high-level guidance for the subsequent Algorithm Decomposer. This component operates without task-specific fine-tuning for its abstract plan generation stage.

\subsection{Algorithm Decomposer}
\label{subsec:gta_decomposer}

Bridging the gap between the high-level plan $\mathcal{A}$ generated previously and the final execution phase, the \textbf{Algorithm Decomposer} refines the abstract strategy into a concise sequence of concrete, manageable sub-steps $\{s_1, s_2, ..., s_k\}$. These sub-steps are specifically crafted for sequential processing by the LLM Executor. Breaking the problem down in this way reduces the load on the Executor, walking its attention incrementally through the algorithmic logic. The transformation performed by the decomposer can be represented as:
\begin{equation}
    \text{Decomposer}: (\mathcal{A}, G_{R^*}) \mapsto \{s_1, s_2, ..., s_k\}
\end{equation}
Each resulting sub-step $s_i$ constitutes a focused natural language instruction that directs the Executor to perform a specific part of the overall algorithm.
To effectively translate conceptual plans into actionable steps, the Algorithm Decomposer is implemented as a dedicated LLM. The specific fine-tuning process designed to optimize its decomposition capabilities is detailed in \autoref{subsec:gta_training_workflow}.

\subsection{Training and Execution Workflow}
\label{subsec:gta_training_workflow}

The GTA framework's components are trained sequentially. The \textbf{Heuristic-Based Selector} is a parameter-free function based on empirical results, and the \textbf{Algorithm Generator} employs a base LLM without specific fine-tuning for its role. The core training focuses on the \textbf{Algorithm Decomposer} and subsequently the \textbf{Learned Selector}.

\textbf{Algorithm Decomposer Training:}
The Algorithm Decomposer LLM is trained in two stages, using the Heuristic-Based Selector for input graph representation ($G_{R^*_{\text{heuristic}}}$).
\textit{1. Supervised Fine-Tuning (SFT):} To establish foundational decomposition skills, the Algorithm Decomposer LLM is initially fine-tuned on expert-like decompositions $D_{\text{expert}}$ (obtained via distillation from a stronger model) for given plans $\mathcal{A}$ and graphs $G_{R^*_{\text{heuristic}}}$. This stage minimizes the SFT loss:
\begin{equation}
    \mathcal{L}_{\text{SFT-Decomp}} = -\mathbb{E}_{(\mathcal{A}, G_{R^*_{\text{heuristic}}}), D_{\text{expert}}} \left[ \log P_{\text{Decomp}}(D_{\text{expert}} | \mathcal{A}, G_{R^*_{\text{heuristic}}}) \right],
\end{equation}
where $P_{\text{Decomp}}$ denotes the probability assigned by the Algorithm Decomposer LLM.
\textit{2. Direct Preference Optimization (DPO):} Following \citet{rafailov2023direct}, the SFT-tuned model is then further refined. Multiple decomposition candidates $\{D_j\}$ are sampled for each problem. After execution by the Executor, these are labeled as ``winning'' ($D_w$, leading to a correct solution) or ``losing'' ($D_l$, leading to an incorrect solution). DPO then optimizes the Algorithm Decomposer LLM using these preference pairs $(D_w, D_l)$ to maximize the likelihood of generating effective decompositions for the Executor. This directly links the Decomposer's output to successful problem execution.

\textbf{Learned Selector Training:}
With the Algorithm Decomposer trained and fixed, the Learned Selector LLM is trained. For each problem instance $P=(G,T)$, the complete GTA pipeline (using the Algorithm Generator, Algorithm Decomposer, and Executor) is executed with each of the four input representations $R \in \{\text{NL, SL, AM, AL}\}$. This yields preferred representations $R_w$ (leading to correct solutions) and dispreferred ones $R_l$. The Learned Selector is then fine-tuned via DPO using these preference pairs $(R_w, R_l)$ for each problem $P$. This optimizes the Selector to choose the representation that maximizes the end-to-end success probability of the GTA pipeline.

\textbf{Execution Phase:}
During inference, the chosen Selector (either Heuristic-Based or Learned Selector) determines the input representation $R^*$. The Algorithm Generator formulates the high-level plan $\mathcal{A}$. The Algorithm Decomposer breaks $\mathcal{A}$ into sub-steps $\{s_i\}$, which are then processed by the base LLM acting as the Executor to derive the final solution.

\section{Evaluating GTA}
\label{sec:experiments}

We evaluate GTA against prompting and agent baselines (\autoref{sec:gta_comparative_performance}), isolate where its gains come from (\autoref{sec:gta_ablation_study}, \autoref{sec:appendix_sft_dpo_dynamics}), and then examine the practical axes that matter for deployment: behavior on larger graphs, transfer to other executors, monetary cost, and failure modes.

\subsection{Setup}
\label{sec:eval_setup}

\textbf{Datasets.}
GTA and all baselines are compared on GT Bench and, to test transfer beyond the training distribution, on the Graph Understanding subset of GraCoRe~\citep{yuan2024gracore} and the complete NLGraph~\citep{wang2023can} dataset. GraCoRe and NLGraph instances are randomly converted to one of the four input modalities used in GT Bench, and each method is evaluated on 1{,}000 instances per dataset.

\textbf{Baselines.} Across all datasets, we compare our proposed GTA against several established methods. These encompass prompting-based techniques: Vanilla prompting, Chain-of-Thought (CoT)~\citep{wei2022chain}, LLM-Debate~\citep{du2023improving}, and Self-Refine~\citep{madaan2023self}; GraphTeam in its reasoning-only configuration~\citep{li2025graphteamfacilitatinglargelanguage}; as well as automated agent frameworks: ADAS~\citep{hu2024automated} , AFlow~\citep{zhang2024aflow} and MaAS~\citep{zhang2025multi}.
Throughout our experiments, ``Vanilla prompting'' refers to using the base LLM executor to solve graph problems directly in a zero-shot setting, without any component of the GTA agentic framework or additional prompting strategies.
To ensure a fair comparison, all baseline methods also utilize the \texttt{Phi-4} model as their executor LLM. Importantly, under all configurations (including GTA), the executor LLM remains frozen and unmodified; GTA's auxiliary components learn to \emph{format and structure} problems for the executor, not the task solutions themselves. This ensures a fair comparison: all methods use the same base model, and only the external reasoning strategy differs.

\textbf{Implementation Details.}
GTA is trained solely on GT Bench and evaluated for generalization on GraCoRe and NLGraph. We use \texttt{Phi-4} for the Learned Selector, Algorithm Generator, Algorithm Decomposer, and Executor to balance performance and computational cost, and distill expert demonstrations for the Decomposer from \texttt{GPT-4o}. Benchmark accuracies are evaluated at a temperature of 0.01.
Further details of the experimental configuration are provided in \autoref{appendix:experimental_details}, and task-level representation results are provided in \autoref{appendix:experimental_results}.

\subsection{Comparative Performance of GTA}
\label{sec:gta_comparative_performance}

We first compare GTA with the prompting strategies and automated agent frameworks introduced above. \autoref{tab:method_comparison} reports average accuracy on GT Bench (Easy and Hard), GraCoRe, and NLGraph.

\needspace{15\baselineskip}
\begin{wraptable}{r}{0.5\textwidth}
\vspace{-1.4em}
\begingroup
\definecolor{headerbg3}{RGB}{242,235,250}
\definecolor{rowone3}{RGB}{249,246,253}
\definecolor{rowtwo3}{RGB}{240,235,247}
\definecolor{gtarow}{RGB}{230,220,245}
\setlength{\arrayrulewidth}{0.5pt}
\arrayrulecolor{gray!50}
\rowcolors{2}{rowone3}{rowtwo3}

\centering
\caption{Average accuracy (\%) of different methods on multiple graph reasoning benchmarks. GT-E = GT Bench (Easy), GT-H = GT Bench (Hard), GCR = GraCoRe, NLG = NLGraph.}
\label{tab:method_comparison}
\vspace{0.5em}
\setlength{\tabcolsep}{5.5pt}
\begin{tabular}{>{\columncolor{headerbg3}}lcccc}
\toprule
\rowcolor{headerbg3}
\textbf{Method} & \textbf{GT-E} & \textbf{GT-H} & \textbf{GCR} & \textbf{NLG} \\
\midrule
Vanilla       & 53.5 & 33.0 & 66.8 & 80.2 \\
CoT          & 52.1 & 34.2 & 67.2 & 81.0 \\
LLM-Debate   & 58.9 & 35.5 & 70.3 & 84.4 \\
Self-Refine  & 56.8 & 34.6 & 68.0 & 83.1 \\
GraphTeam    & 50.1 & 27.2 & 51.6 & 63.2 \\
ADAS        & 52.8 & 32.7 & 67.5 & 80.5 \\
AFlow        & 62.0 & 37.2 & 75.4 & 86.2 \\
MaAS       & 63.2 & 37.6 & 78.8 & 86.0 \\
\midrule
\rowcolor{gtarow}
\textbf{GTA} & \textbf{69.1} & \textbf{41.5} & \textbf{80.5} & \textbf{89.4} \\
\bottomrule
\end{tabular}
\endgroup
\vspace{-1em}
\end{wraptable}

GTA achieves the highest accuracy on every benchmark, reaching 69.1\% on GT-E and 41.5\% on GT-H, ahead of the next best methods, MaAS (63.2\%/37.6\%) and AFlow (62.0\%/37.2\%). Notably, the search-based agent frameworks (ADAS, AFlow, MaAS) optimize the workflow around the same frozen executor, so the comparison isolates what is gained by making the input representation and the algorithmic breakdown explicit targets of optimization rather than searching over generic prompt-and-call structures.

Since GTA is trained solely on GT Bench, its accuracy on GraCoRe (80.5\%) and NLGraph (89.4\%), two benchmarks whose task distributions are not subsets of GT Bench, indicates that adaptive representation selection and structured decomposition transfer to related graph reasoning settings rather than overfitting the training benchmark.

\subsection{Ablation Study of GTA Components}
\label{sec:gta_ablation_study}

\begin{wraptable}{r}{0.55\textwidth}
\vspace{-1.4em}
\begingroup
\definecolor{headerbg4}{RGB}{255,244,234}   
\definecolor{rowone4}{RGB}{255,250,245}     
\definecolor{rowtwo4}{RGB}{250,247,242}     
\definecolor{gtaablation}{RGB}{255,237,217} 
\setlength{\arrayrulewidth}{0.5pt}
\arrayrulecolor{gray!40}
\rowcolors{2}{rowone4}{rowtwo4}
\centering
\caption{Accuracy (\%) on GT Bench, GraCoRe (GCR), and NLGraph (NLG). Configurations: Vanilla (Vanilla Phi-4), GTA-NoSel (w/o Selector), GTA-HeuSel (w/ Heuristic Selector), GTA-NoDec (w/o Generator \& Decomposer), \textbf{GTA} (Our proposed method).}
\label{tab:ablation_study_results}
\vspace{0.5em}
\setlength{\tabcolsep}{6.5pt}
\begin{tabular}{>{\columncolor{headerbg4}}lcccc}
\toprule
\rowcolor{headerbg4}
\textbf{Config.} & \textbf{GT-E} & \textbf{GT-H} & \textbf{GCR} & \textbf{NLG} \\
\midrule
Vanilla       & 53.5 & 33.0 & 66.8 & 80.2 \\
\midrule
GTA-NoSel     & 58.2 & 36.1 & 71.6 & 83.5  \\
GTA-NoDec     & 62.1 & 37.7 & 75.3 & 85.0 \\
\midrule
GTA-HeuSel    & 64.5 & 38.8 & 76.5 & 85.8 \\
\rowcolor{gtaablation}
\textbf{GTA} & \textbf{69.1} & \textbf{41.5} & \textbf{80.5} & \textbf{89.4} \\
\bottomrule
\end{tabular}
\endgroup
\vspace{-1em}
\end{wraptable}

To verify the contribution of GTA's core modules, we compare the full system against a \textbf{Vanilla} baseline (Phi-4 with vanilla prompting) and several simplified variants:
\textbf{GTA-NoSel}: GTA without any Input Representation Selector (fixed default representation).
\textbf{GTA-NoDec}: GTA without the Algorithm Generator and Decomposer (direct execution).
\textbf{GTA-HeuSel}: GTA using the Heuristic-Based Selector instead of the Learned Selector.
Performance is reported in \autoref{tab:ablation_study_results}.

Both modules matter, and neither alone accounts for the gains. Removing representation selection is the single most damaging change: GT-E falls from 69.1\% to 58.2\%, confirming that the input format is not a cosmetic detail but a central factor in whether the executor succeeds. Replacing the Learned Selector with the heuristic lookup costs 4.6 points on GT-E (64.5\% versus 69.1\%), which quantifies how much executor- and task-specific signal the learned policy captures beyond coarse empirical rules. Removing planning and decomposition costs 7.0 points (62.1\% versus 69.1\%), showing that a structured multi-step breakdown helps the executor well beyond direct monolithic solving. The same ordering of configurations holds on GT-H, GraCoRe, and NLGraph, so the two mechanisms contribute complementary improvements that are not tied to a single dataset.

\paragraph{Per-component ablation: Generator versus Decomposer.}
The plan--decompose module instantiates the plan--decompose--execute paradigm widely adopted in prior agent work \citep{wang2023plan,zhou2023least,khot2023decomposed,shen2023hugginggpt}, in which a strategic Generator and a step-level Decomposer play distinct but complementary roles. To isolate their individual contributions, we evaluate two finer-grained variants, averaged over three runs: \textbf{GTA-NoDecomposer} keeps the Generator but removes the Decomposer, and \textbf{GTA-NoGen} keeps the Decomposer but removes the Generator (\autoref{tab:component_ablation}).

\begin{table}[ht]
\centering
\small
\caption{Per-component ablation of the plan--decompose module (mean $\pm$ std over 3 runs). GTA-NoDec removes both modules; GTA-NoDecomposer keeps only the Generator; GTA-NoGen keeps only the Decomposer.}
\label{tab:component_ablation}
\setlength{\tabcolsep}{7pt}
\begin{tabular*}{\textwidth}{@{\extracolsep{\fill}}lcccc@{}}
\toprule
\textbf{Configuration} & \textbf{GT-E (\%)} & \textbf{GT-H (\%)} & \textbf{GCR (\%)} & \textbf{NLG (\%)} \\
\midrule
GTA-NoDec (Selector only)         & $62.1 \pm 0.5$ & $37.7 \pm 0.4$ & $75.3 \pm 0.6$ & $85.0 \pm 0.4$ \\
GTA-NoDecomposer (Generator only) & $64.3 \pm 0.6$ & $38.5 \pm 0.5$ & $76.8 \pm 0.5$ & $86.2 \pm 0.5$ \\
GTA-NoGen (Decomposer only)       & $66.8 \pm 0.5$ & $39.9 \pm 0.4$ & $78.1 \pm 0.5$ & $87.5 \pm 0.4$ \\
\textbf{GTA (full)}               & $\mathbf{69.1 \pm 0.4}$ & $\mathbf{41.5 \pm 0.5}$ & $\mathbf{80.5 \pm 0.3}$ & $\mathbf{89.4 \pm 0.4}$ \\
\bottomrule
\end{tabular*}
\end{table}

Removing the Decomposer (a 4.8-point drop on GT-E) hurts more than removing the Generator (a 2.3-point drop), consistent with the Decomposer supplying the actionable step-level signal that the Executor consumes while the Generator contributes the complementary high-level algorithmic frame. The joint removal (7.0 points below full GTA) is close to the sum of the two independent removals (7.1 points), indicating that the modules play distinct, near-additive roles rather than redundant ones.

\textbf{Selector training for direct execution.} GTA-NoDec reuses the Learned Selector trained with the full pipeline. To measure the effect of this training configuration, we retrain a Selector specifically for the direct Phi-4 Executor, with no Generator or Decomposer (\autoref{tab:selector_direct}). The comparison evaluates how well the Selector's $(G,T)\!\mapsto\!R^*$ mapping transfers from the full pipeline to direct execution.

\begin{table}[ht]
\centering
\small
\caption{A Selector retrained specifically for the direct Executor (Selector-Direct) versus GTA-NoDec, which reuses the full-pipeline Selector (mean $\pm$ std over 3 runs).}
\label{tab:selector_direct}
\setlength{\tabcolsep}{7pt}
\begin{tabular*}{\textwidth}{@{\extracolsep{\fill}}lcccc@{}}
\toprule
\textbf{Configuration} & \textbf{GT-E (\%)} & \textbf{GT-H (\%)} & \textbf{GCR (\%)} & \textbf{NLG (\%)} \\
\midrule
GTA-NoDec (Selector from full pipeline)                   & $62.1 \pm 0.5$ & $37.7 \pm 0.4$ & $75.3 \pm 0.6$ & $85.0 \pm 0.4$ \\
Selector-Direct (Selector retrained for direct Executor)  & $63.8 \pm 0.6$ & $38.4 \pm 0.5$ & $76.2 \pm 0.5$ & $85.6 \pm 0.5$ \\
\textbf{GTA (full)}                                       & $\mathbf{69.1 \pm 0.4}$ & $\mathbf{41.5 \pm 0.5}$ & $\mathbf{80.5 \pm 0.3}$ & $\mathbf{89.4 \pm 0.4}$ \\
\bottomrule
\end{tabular*}
\end{table}

Retraining the Selector for the direct Executor improves accuracy by 1.7 points on GT-E and 0.7 on GT-H over GTA-NoDec. Full GTA still outperforms the retrained Selector-only system by 5.3 points on GT-E and 3.1 on GT-H. Representation selection thus transfers across execution configurations, while the plan--decompose module provides additional gains beyond a Selector optimized for direct execution.

\paragraph{Reliability of the training-free Generator.}
\label{sec:generator_stability}
Because the Algorithm Generator is a prompted, off-the-shelf LLM rather than a fine-tuned module, we measured its stability on a 1{,}000-instance subset spanning all 44 task--structure settings, averaged over three runs. Plan self-consistency, measured by Jaccard overlap over the enumerated reasoning phases at temperature 0.7, is $94.7 \pm 1.2\%$, and algorithmic alignment with the canonical reference algorithms in \autoref{appendix:task_definitions_algorithms}, judged by GPT-4o, is $91.5 \pm 1.8\%$. The GPT-4o alignment score is an auxiliary judge-based diagnostic; LLM-judge bias has been studied mechanistically by \citet{xu2026unfairjudge}. Other evaluations document systematic judgment biases \citep{ye2025justice}, preference leakage when generators and judges are related \citep{li2026prefleakage}, and susceptibility to optimized prompt injection \citep{shi2024judgedeceiver}. GT Bench answer accuracy is instead evaluated against algorithmically verified ground truth. These diagnostics indicate that the Generator produces stable, well-aligned high-level plans for these classical graph problems, which justifies keeping the Generator training-free and concentrating optimization on the Decomposer and the Selector.

\subsection{SFT--DPO Training Dynamics of the Learned Components}
\label{sec:appendix_sft_dpo_dynamics}

We employ a two-stage pipeline for the Algorithm Decomposer and the Learned Selector: (i) \textbf{SFT} on expert demonstrations to establish a broad, modality-agnostic foundation, followed by (ii) \textbf{DPO} to directly optimize for end-to-end task success beyond imitation-style learning.

\textbf{Selector choice during Decomposer training.} We compare DPO training with the Heuristic-Based Selector against a \emph{Random Selector} that samples all input modalities uniformly. This comparison measures the effect of the training-time representation policy on the resulting Decomposer. Both configurations improve over the Vanilla baseline, with the heuristic configuration achieving higher mean accuracy across the four evaluation sets:

\begin{table}[ht]
\centering
\small
\caption{Effect of selector choice during DPO on Decomposer performance (mean $\pm$ std).}
\label{tab:dpo_selector_bias}
\begin{tabular*}{\textwidth}{@{\extracolsep{\fill}}lcccc@{}}
\toprule
\textbf{Decomposer trained with} & \textbf{GT-E (\%)} & \textbf{GT-H (\%)} & \textbf{GCR (\%)} & \textbf{NLG (\%)} \\
\midrule
Vanilla Baseline & $53.5 \pm 0.2$ & $33.0 \pm 0.3$ & $66.8 \pm 0.1$ & $80.2 \pm 0.2$ \\
Heuristic Selector (DPO) & $\mathbf{69.1 \pm 0.4}$ & $\mathbf{41.5 \pm 0.5}$ & $\mathbf{80.5 \pm 0.3}$ & $\mathbf{89.4 \pm 0.4}$ \\
Random Selector (DPO) & $68.7 \pm 0.7$ & $40.9 \pm 0.6$ & $79.8 \pm 0.9$ & $87.0 \pm 0.5$ \\
\bottomrule
\end{tabular*}
\end{table}

\textbf{Module-specific DPO gains.} We apply DPO to two modules:

1) \emph{Learned Selector.} DPO enables the selector to prefer the input representation that most likely yields a correct answer. As shown in \autoref{tab:ablation_study_results}, the full GTA (with the Learned Selector) substantially outperforms the heuristic variant (GTA-HeuSel), confirming the value of DPO-guided representation selection.

2) \emph{Algorithm Decomposer.} To quantify the decomposer-side gains, we compare SFT-only versus SFT+DPO. Averaged over 3 runs:

\begin{table}[ht]
\centering
\small
\caption{Decomposer ablation: SFT-only vs.\ SFT{+}DPO (mean $\pm$ std).}
\label{tab:sft_vs_sftdpo}
\begin{tabular*}{\textwidth}{@{\extracolsep{\fill}}lcccc@{}}
\toprule
\textbf{Config. (Decomposer trained with\ldots)} & \textbf{GT-E (\%)} & \textbf{GT-H (\%)} & \textbf{GCR (\%)} & \textbf{NLG (\%)} \\
\midrule
SFT Only & $65.1 \pm 1.1$ & $37.2 \pm 0.8$ & $71.5 \pm 0.9$ & $80.3 \pm 0.7$ \\
SFT + DPO (Ours) & $\mathbf{69.1 \pm 0.4}$ & $\mathbf{41.5 \pm 0.5}$ & $\mathbf{80.5 \pm 0.3}$ & $\mathbf{89.4 \pm 0.4}$ \\
\bottomrule
\end{tabular*}
\end{table}

The SFT stage supplies demonstrations across tasks and modalities, while DPO optimizes for task success. DPO-trained Decomposers improve over the Vanilla baseline under both training-time selector policies (Table~\ref{tab:dpo_selector_bias}), and SFT{+}DPO outperforms SFT-only training (Table~\ref{tab:sft_vs_sftdpo}). These results complement the Learned Selector ablation (\autoref{tab:ablation_study_results}) and identify the contributions of the two trained components to GTA's end-to-end performance.

\subsection{Scalability Analysis with Increasing Graph Size}
\label{sec:appendix_scalability}

\begin{figure}[ht]
    \centering
    \includegraphics[width=0.8\linewidth]{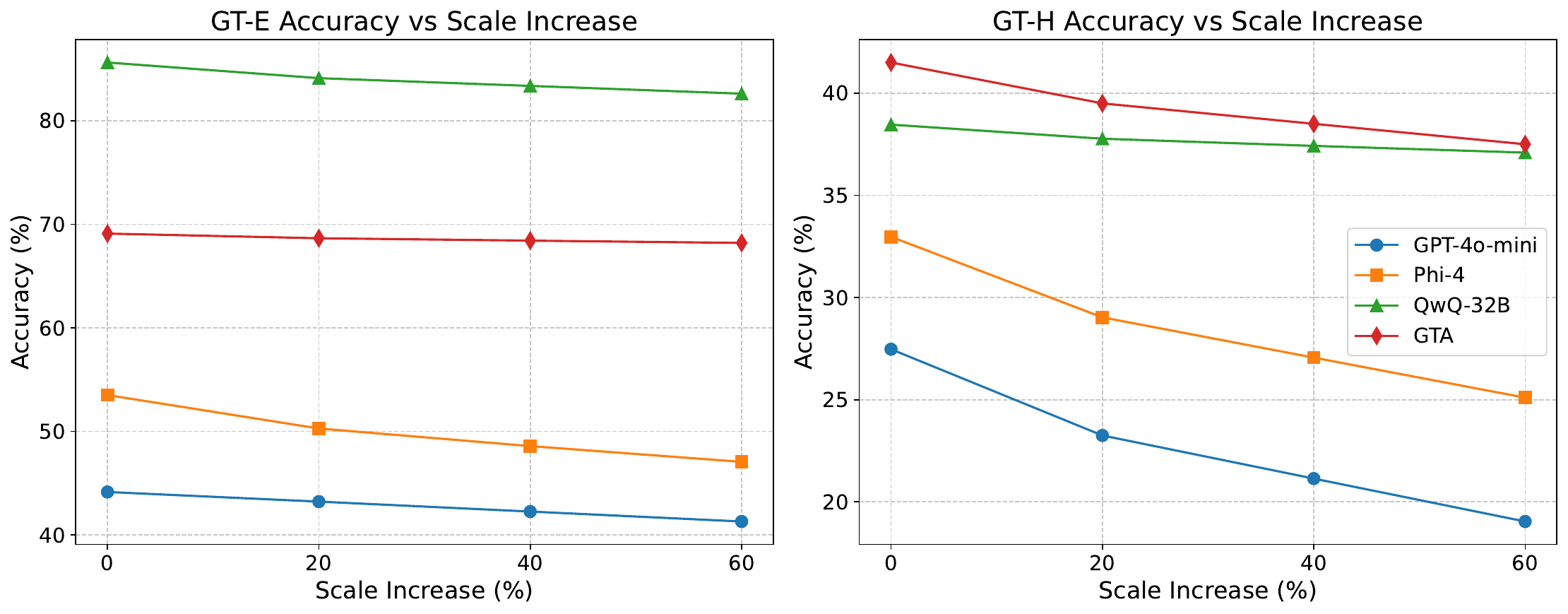}
    \caption{Performance comparison of different models (Phi-4, GPT-4o-mini, QwQ-32B) and the GTA framework on GT Bench Easy (GT-E) and Hard (GT-H) subsets as the number of graph scale increases.}
    \label{fig:accuracy_comparison}
\end{figure}

A key advantage of our automated generation process for GT Bench is the flexibility to adjust graph size (number of nodes and edges), enabling the creation of customized datasets to study LLM scalability on graph reasoning tasks. To investigate the impact of graph scale on performance, we generated variants of GT Bench tasks with increasing numbers of nodes. \autoref{fig:accuracy_comparison} illustrates the performance trends of selected models (\texttt{Phi-4}, \texttt{GPT-4o-mini}, \texttt{QwQ-32B}) and our GTA framework (using \texttt{Phi-4} as the base) as graph size increases, separately for Easy (GT-E) and Hard (GT-H) subsets.

On the GT-E subset, the accuracy of the smaller base models, \texttt{Phi-4} and \texttt{GPT-4o-mini}, declines steadily as the node count grows, while GTA stays comparatively flat across the tested sizes. For these simpler tasks, adaptive representation plus decomposition absorbs most of the difficulty that scale adds.

The picture changes on GT-H. As graphs grow, \texttt{Phi-4} and \texttt{GPT-4o-mini} fall off sharply, and GTA declines as well: its drop is steeper than that of a strong standalone reasoner such as \texttt{QwQ-32B}, but far gentler than that of its own base model, which it outperforms at every size tested. The framework mitigates, rather than eliminates, the difficulty of scaling hard combinatorial tasks; on the largest hard instances the executor's intrinsic capability reasserts itself as the binding constraint.

In short, GTA does not remove the difficulty of large hard instances, but for the same base model it consistently turns a steep degradation curve into a gentler one.

\subsection{Generality across Executor Models}
\label{sec:appendix_executor_choice}

While the core GTA components (Learned Selector, Algorithm Decomposer) were fine-tuned using \texttt{Phi-4}, it is important to understand how the framework performs when a different LLM acts as the final Executor. To investigate this, we evaluated the performance of GTA using \texttt{GPT-4o-mini}, \texttt{GPT-4o}, and \texttt{DeepSeek-R1} as executors, comparing their performance within the GTA framework against their respective vanilla (direct prompting) results. The outcomes are presented in \autoref{tab:gta_executor_comparison}.

\begin{table}[htbp]
    \centering
    \small
    \caption{Accuracy (\%) comparison of different models with Vanilla prompting vs. integrated into the GTA framework as the Executor. GTA components (Selector, Decomposer) were trained using Phi-4.}
    \label{tab:gta_executor_comparison}
    \begin{tabular*}{\textwidth}{@{\extracolsep{\fill}}lcccc@{}}
        \toprule
        \textbf{Model Configuration} & \textbf{GT-E (\%)} & \textbf{GT-H (\%)} & \textbf{GCR (\%)} & \textbf{NLG (\%)} \\
        \midrule
        GPT-4o-mini (Vanilla)     & 44.2 & 27.5 & 56.2 & 70.3 \\
        GPT-4o-mini (GTA Executor)  & \textbf{53.1} & \textbf{34.2} & \textbf{63.5} & \textbf{76.6}  \\
        \midrule
        Phi-4 (Vanilla)           & 53.5 & 33.0 & 66.8 & 80.2 \\
        Phi-4 (GTA Executor)        & \textbf{69.1} & \textbf{41.5} & \textbf{80.5} & \textbf{89.4} \\
        \midrule
        GPT-4o (Vanilla)    & 60.5 & 28.9 & 70.2 & 81.4 \\
        GPT-4o (GTA Executor) & \textbf{68.8} & \textbf{36.9} & \textbf{78.8} & \textbf{88.2} \\
        \midrule
        DeepSeek-R1 (Vanilla)    & 96.3 & 64.1 & 97.6 & 99.0 \\
        DeepSeek-R1 (GTA Executor) & 96.4 & \textbf{72.2} & 97.6 & 98.8 \\
        \bottomrule
    \end{tabular*}
\end{table}

GTA transfers across executors despite its components having been tuned with \texttt{Phi-4}. \texttt{GPT-4o-mini} and \texttt{GPT-4o} gain between 6 and 9 points on every benchmark; \texttt{GPT-4o}, for instance, rises from 60.5\% to 68.8\% on GT-E and from 28.9\% to 36.9\% on GT-H. The \texttt{DeepSeek-R1} row shows how these gains behave at the top of the capability range: where the executor is already near ceiling (GT-E, GraCoRe, NLGraph), the observed changes range from $-0.2$ to $+0.1$ percentage points, while on the unsaturated GT-H split it adds 8.1 points (64.1\% to 72.2\%). The largest gain for this executor occurs on the benchmark split with the most remaining headroom.

\subsection{Inference Cost Analysis}
\label{sec:appendix_inference_cost}

Beyond accuracy, inference cost decides whether a reasoning framework is practical to deploy. To compare the efficiency of GTA with the baselines, we measured token consumption and calculated the associated costs when running each method on a representative sample of 600 instances drawn from the GT Bench dataset (combining both Easy and Hard tasks). For all methods, \texttt{Phi-4} was used as the base LLM to ensure a fair comparison of the overhead introduced by each framework. Costs are calculated from the reported token counts at rates of \$0.07 per million input tokens and \$0.14 per million output tokens. \autoref{tab:inference_cost} summarizes these findings.

\begin{table}[ht]
\centering
\caption{Inference Cost and Token Consumption (in Millions) for running 600 instances (mixed Easy/Hard) from GT Bench using Phi-4 as the base LLM. Pricing: \$0.07/M input, \$0.14/M output.}
\label{tab:inference_cost}
\begin{tabular*}{\textwidth}{@{\extracolsep{\fill}}lccc@{}}
\toprule
\textbf{Method} & \textbf{Input Tokens (M)} & \textbf{Output Tokens (M)} & \textbf{Calculated Cost (USD)} \\
\midrule
Phi-4 (Vanilla) & 1.75 & 0.53 & \$0.20 \\
LLM-Debate    & 64.51 & 19.81 & \$7.29 \\
Self-Refine   & 34.87 & 10.71 & \$3.94 \\
AFlow         & 17.44 & 5.65 & \$2.01 \\
MaAS          & 8.11 & 2.49 & \$0.92 \\
GTA (ours)    & 9.85 & 3.02 & \$1.11 \\
\bottomrule
\end{tabular*}

\end{table}

Cost differences across methods are large. Multi-round methods are the most expensive: LLM-Debate consumes over thirty times the tokens of vanilla prompting, and Self-Refine and AFlow also add substantial overhead. GTA's overhead is instead bounded by construction. Each instance passes through the Selector, Generator, Decomposer, and Executor once, so token usage grows by a small constant factor over vanilla prompting (roughly 5--6$\times$ here) rather than with the number of interaction rounds. GTA therefore delivers the highest accuracy in \autoref{tab:method_comparison} at \$1.11 per 600 instances, below every method in \autoref{tab:inference_cost} except MaAS and vanilla prompting itself.

\paragraph{One-time training and data-generation cost.}
The inference figures above exclude the one-time overhead of building GTA, which we itemize in \autoref{tab:training_cost}. Distilling decomposition demonstrations from \texttt{GPT-4o} for SFT consumes roughly 5M tokens (about \$28), and DPO fine-tuning together with Selector preference-pair generation runs for 40 hours on two A6000 GPUs (about \$64), for a total of roughly \$92.

\begin{table}[ht]
\centering
\small
\caption{One-time training and data-generation cost of GTA. This overhead is incurred once and amortizes across all subsequent inferences.}
\label{tab:training_cost}
\setlength{\tabcolsep}{7pt}
\resizebox{\textwidth}{!}{%
\begin{tabular}{lcc}
\toprule
\textbf{Component} & \textbf{Quantity} & \textbf{Subtotal (USD)} \\
\midrule
GPT-4o distillation for SFT (1{,}000 instances) & $\sim$5M tokens (3M in / 2M out) & $\sim$\$28 \\
DPO fine-tuning + Selector preference-pair generation (2$\times$A6000, 40h) & 80 GPU-hours & $\sim$\$64 \\
\midrule
\textbf{Total one-time cost} & & $\boldsymbol{\approx}$\textbf{\$92} \\
\bottomrule
\end{tabular}%
}
\end{table}

The \$92 one-time cost adds \$0.0092 per instance when spread over 10{,}000 inferences and \$0.00092 per instance over 100{,}000 inferences. For comparison, the token counts in \autoref{tab:inference_cost} yield an inference cost of approximately \$0.00185 per instance, or \$1.85 per 1{,}000 instances. Total cost therefore depends on evaluation volume as well as inference token consumption.

\subsection{Failure-Mode and Truncation Analysis}
\label{sec:truncation}

We measure context overflow, output truncation, and answer-extraction failures to assess their contribution to the reported error rates. All evaluations use a maximum generation length of 8{,}192 tokens, consistent across models. \autoref{tab:truncation} reports these statistics across all 10{,}000 evaluation instances, averaged over three runs.

\begin{table}[ht]
\centering
\small
\caption{Input-overflow, output-truncation, and answer-extraction rates across all 10{,}000 evaluation instances (mean $\pm$ std over 3 runs). Input overflow is eliminated by design through the prompt-fit budget of GT Bench.}
\label{tab:truncation}
\setlength{\tabcolsep}{8pt}
\begin{tabular*}{\textwidth}{@{\extracolsep{\fill}}lccc@{}}
\toprule
\textbf{Metric} & \textbf{Overall} & \textbf{GT-E (Easy)} & \textbf{GT-H (Hard)} \\
\midrule
Input exceeds context window   & $0.0\%$ & $0.0\%$ & $0.0\%$ \\
Output truncated at max length & $1.2 \pm 0.3\%$ & $0.5 \pm 0.2\%$ & $2.1 \pm 0.4\%$ \\
Answer extraction failure      & $0.8 \pm 0.2\%$ & $0.5 \pm 0.1\%$ & $1.2 \pm 0.3\%$ \\
\bottomrule
\end{tabular*}
\end{table}

The 0.0\% input-overflow rate is by design: the prompt-fit budget of \autoref{sec:graphbench_dataset} calibrates node ranges so that all four representations fit within typical context windows, with tighter ranges for dense graphs where the edge count grows as $O(V^2)$. Output truncation is low (1.2\% overall, 2.1\% on Hard) and cannot account for the observed accuracy gaps: the gap between GTA-NoDec and full GTA alone is 7.0 points on GT-E and 3.8 on GT-H, far larger than the 1--2\% truncation rate. Answer-extraction failures are similarly rare (0.8\% overall). The performance differences we report therefore reflect reasoning ability rather than formatting or truncation artifacts.

\section{Conclusion}
We introduced GT Bench, a benchmark of 24 graph problems in 44 task--structure settings with over 100{,}000 examples across four input representations, and used it to characterize how LLMs reason over graphs in language. The evaluation yields a consistent picture: accuracy depends on the input representation in ways that track task, graph structure, and model; the sensitivity persists, attenuated, in the strongest reasoning models; and hard multi-step tasks remain far from solved by scale alone. The Graph Theory Agent turns these observations into a method. A preference-trained Selector picks the representation per instance and a Generator--Decomposer pair converts the task into steps a frozen executor can follow, yielding consistent gains over prompting and agent baselines at bounded cost, transfer to benchmarks and executors unseen in training, and a gentler degradation curve as graphs grow (\autoref{sec:appendix_scalability}). Together, GT Bench and GTA provide a benchmark and an adaptive reasoning framework for measuring and improving language-based algorithmic graph reasoning.

\clearpage

\phantomsection
\addcontentsline{toc}{section}{References}
\bibliography{main}
\bibliographystyle{gta_arxiv}
\clearpage

\appendix
\section{Task Definitions and Solution Algorithms in GT Bench}
\label{appendix:task_definitions_algorithms}

This section provides detailed definitions for the graph-theoretic tasks included in GT Bench, along with the canonical algorithms used to generate their ground-truth solutions.
We begin by defining fundamental graph-theoretic concepts used throughout this section. A \textbf{graph} $G$ is represented as a pair $(V,E)$, where $V$ is a set of \textbf{vertices} (also called nodes) and $E$ is a set of \textbf{edges} connecting pairs of vertices. Graphs can be \textbf{directed} (edges have a direction from one vertex to another) or \textbf{undirected} (edges have no intrinsic direction). Edges may also have associated \textbf{weights} or \textbf{capacities}. A \textbf{path} is a sequence of vertices such that from each of its vertices there is an edge to the next vertex in the sequence. A \textbf{simple path} is a path where all vertices are distinct. A \textbf{cycle} is a path that starts and ends at the same vertex. A \textbf{simple cycle} is a cycle where all intermediate vertices are distinct.

\paragraph{Connectivity.}
Definition: This task requires determining whether a path exists between two specified vertices, say $u$ and $v$, in a given graph $G=(V,E)$.
Solution Algorithm: The ground truth is established by performing a Breadth-First Search (BFS) starting from vertex $u$. If vertex $v$ is visited during the traversal, the vertices are deemed connected; otherwise, they are not.

\paragraph{Bipartiteness Check.}
Definition: This task aims to determine if the vertices of a graph $G=(V,E)$ can be partitioned into two disjoint and independent sets, $U$ and $W$, such that every edge in $E$ connects a vertex in $U$ to one in $W$.
Solution Algorithm: A graph coloring approach using two colors is implemented via BFS. Starting from an arbitrary uncolored vertex and assigning it a color, its neighbors are assigned the other color, and this process continues. If an edge is found to connect two vertices of the same color, the graph is not bipartite; otherwise, it is.

\paragraph{Minimum Cycle Length.}
Definition: For an unweighted graph $G=(V,E)$, this task is to find the length (i.e., the number of edges) of the shortest simple cycle.
Solution Algorithm: The length of the shortest cycle is found by iterating through each vertex $s \in V$. For each $s$, a BFS is initiated. If the BFS encounters a previously visited vertex $t$ (other than the immediate parent of the current vertex in the BFS tree for $s$), a cycle is found. The length of this cycle is $\text{depth}(s_{\text{current}}) + \text{depth}(t) + 1$. The minimum such length over all starting vertices $s$ and all detected cycles forms the minimum cycle length.

\paragraph{Maximum Clique Size.}
Definition: A clique in an undirected graph $G=(V,E)$ is a subset of vertices such that every two distinct vertices in the clique are adjacent. This task requires finding the number of vertices in the largest such subset.
Solution Algorithm: Given that finding the maximum clique is NP-hard, a backtracking-based brute-force search algorithm is employed. This algorithm systematically explores all possible subsets of vertices, checking if each forms a clique, and keeps track of the largest clique found.

\paragraph{Maximum Independent Set Size.}
Definition: An independent set in an undirected graph $G=(V,E)$ is a subset of vertices such that no two vertices in the subset are adjacent. This task asks for the size of the largest such set.
Solution Algorithm: This problem is NP-hard. Similar to the maximum clique problem, a backtracking-based brute-force search algorithm is used. It explores all subsets of vertices, verifies the independent set property, and records the size of the largest valid independent set encountered.

\paragraph{Eulerian Path.}
Definition: An Eulerian path in a graph is a trail that visits every edge exactly once. This task determines whether such a path exists.
Solution Algorithm: For an undirected graph $G$, an Eulerian path exists if and only if $G$ is connected (considering only non-isolated vertices) and has exactly zero or two vertices of odd degree. Connectivity is checked using BFS. The degrees of all vertices are computed, and these conditions are verified.

\paragraph{Eulerian Circuit.}
Definition: An Eulerian circuit is an Eulerian path that starts and ends on the same vertex. This task determines if such a circuit exists.
Solution Algorithm: For an undirected graph $G$, an Eulerian circuit exists if and only if all non-isolated vertices belong to one connected component and every vertex has even degree. Connectivity among non-isolated vertices is checked using BFS.

\paragraph{Hamiltonian Path.}
Definition: A Hamiltonian path is a path in an undirected or directed graph that visits each vertex exactly once. This task is to determine if such a path exists.
Solution Algorithm: This problem is NP-complete. A backtracking-based brute-force search algorithm is employed. The algorithm attempts to build a path vertex by vertex, ensuring each vertex is visited exactly once, exploring all valid sequences.

\paragraph{Hamiltonian Circuit.}
Definition: A Hamiltonian circuit is a Hamiltonian path that is a cycle.
Solution Algorithm: This problem is NP-complete. A backtracking-based brute-force search, similar to that for Hamiltonian paths, is used, with the additional constraint that the path must form a cycle by connecting the last vertex in the path back to the first.

\paragraph{Biconnected Components Count.}
Definition: A biconnected component of a graph is a maximal biconnected subgraph. A graph is biconnected if it remains connected if any single vertex is removed. This task counts the number of biconnected components.
Solution Algorithm: The ground truth is found using a Depth-First Search (DFS)-based algorithm that identifies articulation points and bridges. Edges are then grouped into biconnected components based on these findings.

\paragraph{Bridge Count.}
Definition: A bridge in an undirected graph is an edge whose removal increases the number of connected components. This task requires counting the number of such bridges.
Solution Algorithm: A DFS-based algorithm is used. During the DFS, discovery times ($disc[u]$) and low-link values ($low[u]$) are maintained. An edge $(u,v)$ where $v$ is a child of $u$ in the DFS tree is a bridge if $low[v] > disc[u]$.

\paragraph{Triangle Count.}
Definition: This task is to count the number of simple cycles of length 3 (triangles) in an undirected graph.
Solution Algorithm: The number of triangles is computed by iterating through all unique triples of vertices $(u,v,w)$ and checking if edges $(u,v)$, $(v,w)$, and $(w,u)$ all exist in the graph.

\paragraph{Cycle Count.}
Definition: This task requires counting the total number of simple cycles in the graph.
Solution Algorithm: For the small graphs specified for this task in GT Bench (6-8 nodes), a DFS-based backtracking search is implemented to explicitly enumerate and count all elementary circuits.

\paragraph{Spanning Tree Count.}
Definition: This task is to count the number of distinct spanning trees in a given connected undirected graph.
Solution Algorithm: Kirchhoff's Matrix Tree Theorem is applied. The Laplacian matrix $L$ of the graph is constructed ($L = D - A$, where $D$ is the degree matrix and $A$ is the adjacency matrix). Any cofactor of $L$ gives the number of spanning trees.

\paragraph{Shortest Path Length.}
Definition: This task involves computing the length of a shortest path between two specified vertices $s$ and $t$.
Solution Algorithm: For unweighted graphs, Breadth-First Search (BFS) starting from $s$ is used. For weighted graphs with non-negative edge weights, Dijkstra's algorithm is employed.

\paragraph{Minimum Spanning Tree (MST) Weight.}
Definition: Given a connected, undirected, and edge-weighted graph, this task requires computing the total weight of a Minimum Spanning Tree (MST).
Solution Algorithm: Kruskal's algorithm is used to find an MST. The sum of the weights of the edges in the found MST is the result.

\paragraph{Second Minimum Spanning Tree (SMST) Weight.}
Definition: This task asks for the minimum total weight among spanning trees whose weight is strictly greater than the MST weight.
Solution Algorithm: First, an MST $T$ is found using Kruskal's algorithm. For each non-tree edge $(u,v)$, adding it to $T$ creates a cycle. Among the edges of $T$ on this cycle with weight strictly smaller than that of $(u,v)$, the maximum-weight edge $(x,y)$ is selected; if no such edge exists, this candidate is skipped. Replacing $(x,y)$ with $(u,v)$ yields a strictly heavier spanning tree $T'$. The SMST weight is the minimum weight over these candidates.

\paragraph{Tree Diameter.}
Definition: The diameter of a tree is the length of the longest shortest path between any pair of nodes.
Solution Algorithm: This is found using two BFS traversals:
1. Start a BFS from an arbitrary node $s$ to find the node $u$ farthest from $s$.
2. Start another BFS from $u$ to find the node $v$ farthest from $u$.
The distance between $u$ and $v$ is the diameter.

\paragraph{Tree Centroid.}
Definition: A centroid of a tree is a node $c$ such that removing $c$ divides the tree into components each with at most $|V|/2$ nodes. The task is to find such a centroid (smallest index if multiple exist).
Solution Algorithm: A DFS traversal computes subtree sizes. A node $v$ is a centroid if the size of each subtree rooted at a child of $v$, and the size of the remaining tree if $v$ is removed, are all $\le |V|/2$. The centroid with the smallest index is selected.

\paragraph{Lowest Common Ancestor (LCA).}
Definition: Given a rooted tree (node 1 as root) and two nodes $u$ and $v$, the LCA is the deepest node that is an ancestor of both $u$ and $v$.
Solution Algorithm: Depths and parent pointers are computed via DFS. To find LCA($u,v$):
1. Bring $u$ and $v$ to the same depth by moving the deeper node up.
2. If $u=v$, it is the LCA.
3. Else, move both $u$ and $v$ upwards simultaneously until they meet. This meeting point is the LCA.

\paragraph{Tree Maximum Independent Set Size.}
Definition: For a given tree, find the size of a maximum independent set.
Solution Algorithm: Dynamic programming on trees is used. For each node $u$, $dp[u][\text{1}]$ (size of max independent set in subtree of $u$, including $u$) and $dp[u][\text{0}]$ (size, excluding $u$) are computed in a post-order traversal.
$dp[u][\text{1}] = 1 + \sum_{c \in \text{children}(u)} dp[c][\text{0}]$
$dp[u][\text{0}] = \sum_{c \in \text{children}(u)} \max(dp[c][\text{1}], dp[c][\text{0}])$
The answer is $\max(dp[\text{root}][\text{1}], dp[\text{root}][\text{0}])$.

\paragraph{Maximum Flow.}
Definition: Given a flow network with edge capacities, a source $s$, and a sink $t$, find the maximum flow from $s$ to $t$.
Solution Algorithm: The Edmonds-Karp algorithm is used. It iteratively finds augmenting paths from $s$ to $t$ in the residual graph using BFS and pushes flow along these paths until no more exist.

\paragraph{Minimum Cut Capacity.}
Definition: In a flow network, an $s-t$ cut partitions vertices into $S$ (containing $s$) and $T$ (containing $t$). The cut capacity is the sum of capacities of edges from $S$ to $T$. This task asks for the minimum such capacity.
Solution Algorithm: By the Max-Flow Min-Cut Theorem, this value is equal to the maximum flow from $s$ to $t$. Thus, the Edmonds-Karp algorithm is used to compute the maximum flow, which gives the minimum cut capacity.

\paragraph{Minimum-Cost Maximum-Flow Value.}
Definition: In a flow network with edge capacities and costs per unit of flow, find a flow that maximizes total flow and, among such flows, minimizes total cost. The task asks for this minimum cost.
Solution Algorithm: A successive shortest path algorithm using costs as edge lengths in the residual graph is employed. It repeatedly finds the shortest (minimum cost) augmenting path from $s$ to $t$ (using Bellman-Ford due to potential negative costs in residual graphs after flow augmentation) and pushes flow. This continues until no more augmenting paths exist or a pre-calculated maximum flow value is reached. The total cost is accumulated.

\section{Experimental Details}
\label{appendix:experimental_details}

This section provides further details on the experimental setup, including the models utilized, configurations for baseline methods, and specific parameters for training and evaluation.

\subsection{Models}
A variety of LLMs were employed in our experiments, both for direct evaluation and as components within the GTA framework and baseline methods. \autoref{tab:all_models} lists these models with their versions, originating organizations, licenses, and roles in our study (evaluation or fine-tuning).

\definecolor{headerbg}{RGB}{255, 183, 77}
\definecolor{rowgray}{RGB}{255, 236, 209}
\definecolor{rowblue}{RGB}{255, 247, 230}
\definecolor{darkgreen}{RGB}{0, 150, 0}
\begin{table*}[htbp]
\centering
\small
\renewcommand{\arraystretch}{1}
\setlength{\tabcolsep}{4pt}
\caption{Models used in our experiments along with their versions, organizations, licenses, and purposes. \textit{Eval}: Model used for evaluation; \textit{FT}: Model used for fine-tuning.}
\rowcolors{2}{rowblue}{rowgray}
\label{tab:all_models}
\begin{tabularx}{\textwidth}{l>{\raggedright\arraybackslash}Xcccc}
    \toprule[1.5pt]
    \rowcolor{headerbg}
    \textcolor{white}{\textbf{Model}} &
    \textcolor{white}{\textbf{Version}} &
    \textcolor{white}{\textbf{Organization}} &
    \textcolor{white}{\textbf{License}} &
    \textcolor{white}{\textbf{Eval}} &
    \textcolor{white}{\textbf{FT}} \\
    \midrule[0.8pt]
    Phi-4      & Phi-4        & Microsoft   & MIT                    & \textcolor{darkgreen}{\checkmark} & \textcolor{darkgreen}{\checkmark} \\
    GPT-4o-mini       & gpt-4o-mini-2024-07-18       & OpenAI      & Proprietary            & \textcolor{darkgreen}{\checkmark} & \\
    GPT-4o            & gpt-4o-2024-08-06            & OpenAI      & Proprietary            & \textcolor{darkgreen}{\checkmark} & \\
    Llama-3.1-8B      & Meta-Llama-3.1-8B-Instruct   & Meta        & Llama 3.1 Community    & \textcolor{darkgreen}{\checkmark} &  \\
    Llama-3.3-70B     & Meta-Llama-3.3-70B-Instruct  & Meta        & Llama-3.3    & \textcolor{darkgreen}{\checkmark} & \\
    QwQ       & QwQ-32B         & Alibaba     & Apache 2.0           & \textcolor{darkgreen}{\checkmark} & \\
    o3-mini           & o3-mini-2025-01-31           & OpenAI      & Proprietary            & \textcolor{darkgreen}{\checkmark} & \\
    Deepseek-R1      & DeepSeek-R1        & DeepSeek   & MIT                    & \textcolor{darkgreen}{\checkmark} &  \\
    \bottomrule[1.5pt]
\end{tabularx}
\end{table*}

When these models were evaluated directly (i.e., not as part of GTA or other agentic frameworks), we employed Vanilla prompting in a zero-shot setting to assess their baseline graph reasoning capabilities.

\subsection{Baselines}
\label{appendix:baselines}
For all evaluated baseline methods (Vanilla prompting, Chain-of-Thought (CoT)~\citep{wei2022chain}, LLM-Debate~\citep{du2023improving}, Self-Refine~\citep{madaan2023self}, ADAS~\citep{hu2024automated}, AFlow~\citep{zhang2024aflow}, and MaAS~\citep{zhang2025multi}), we standardized the executor LLM to be \texttt{Phi-4}. This was done to ensure a fair comparison of the reasoning strategies or frameworks themselves, isolating their impact from variations in the underlying model's raw capabilities. For the automated agent frameworks (ADAS, AFlow, MaAS), the respective planner or optimizer components were configured following the specifications and recommendations detailed in their original publications.

In addition, we include two graph-specific baselines for context:

\textbf{GraphTeam (Reasoning-Only).}
Following \citet{li2025graphteamfacilitatinglargelanguage}, we ablate the original multi-agent pipeline to a language-only setting by disabling search/coding/tools and external code execution. The executor remains \texttt{Phi-4} for consistency with other agentic baselines, so any improvement reflects coordination/reasoning rather than tool use.

\textbf{GraphWiz-DPO.}
We evaluate the released \emph{GraphWiz-DPO} checkpoint \citep{chen2024graphwizinstructionfollowinglanguagemodel} in its prescribed instruction format and report results alongside a vanilla prompt and our GTA:

\begin{table}[h]
\centering
\small
\setlength{\tabcolsep}{6pt}
\begin{tabular}{lcc}
\toprule
\textbf{Config} & \textbf{GT-E (\%)} & \textbf{GT-H (\%)} \\
\midrule
Vanilla & 30.2 & 15.4 \\
GraphWiz-DPO & 32.5 & 16.6 \\
GTA (LLaMA 2-13B) & 50.8 & 25.7 \\
\bottomrule
\end{tabular}
\end{table}

Other graph systems use different task or execution interfaces. \textbf{Graph Agent}~\citep{wang2023graph} targets node classification and link prediction, while \textbf{GraphAgent-Reasoner}~\citep{hu2024scalableaccurategraphreasoning} distributes computation across node-level agents. \textbf{GraphLLM}~\citep{chai2023graphllmboostinggraphreasoning} learns graph-enhanced prefixes for a frozen LLM. \textbf{GraphThought}~\citep{huang2025graphthoughtgraphcombinatorialoptimization} and \textbf{GUNDAM}~\citep{ouyang2024gundamaligninglargelanguage} train graph-specialized models, and \textbf{GCoder}~\citep{zhang2024gcoderimprovinglargelanguage} solves graph problems through generated code. These differences define the scope of the comparisons above; they do not establish a general ranking of these systems.

Overall, these choices align with our objective: GTA improves performance \emph{by reasoning} (via adaptive representation selection and a lightweight plan–decompose–execute pipeline) without modifying the executor or relying on code execution.

\subsection{Training and Evaluation Parameters}

\textbf{GTA Component Training:}
The fine-tuning of GTA's components involved specific datasets and hyperparameters.
For the \textbf{Algorithm Decomposer}, the Supervised Fine-Tuning (SFT) stage utilized a dataset of 1,000 instances where decomposition steps were distilled from \texttt{GPT-4o}. Subsequently, for Direct Preference Optimization (DPO), 2,000 preference pairs (winning vs. losing decompositions) were used. It is important to note that, given the performance characteristics of the \texttt{Phi-4} model used as the base, we constrained the number of decomposed sub-tasks, $k$, for the Algorithm Decomposer to be no more than 3. Our empirical observations indicated that when $k$ exceeded 3, the computational overhead increased while the overall accuracy of GTA tended to decrease.
For training the \textbf{Learned Selector}, an initial SFT phase was conducted using 1,000 data instances. This was followed by DPO, where preference pairs were generated based on the end-to-end success of the GTA pipeline with different input representations.

Common hyperparameters for all fine-tuning experiments were as follows: models were trained for 2 epochs with a learning rate of 5.0e-6. We employed a cosine learning rate scheduler with a warmup ratio of 0.1. The per-device training batch size was set to 1, with gradient accumulation performed over 8 steps. Training was conducted using bf16 precision to optimize memory and speed. When generating data for SFT or sampling candidates for DPO, a temperature of 1.0 was used to encourage diversity. All training experiments were conducted on 2 NVIDIA A6000 GPUs over a period of 40 hours.

\textbf{Evaluation Settings:}
For benchmark accuracy evaluation, including direct model assessments and runs of the baseline methods and GTA, the inference temperature was set to 0.01. This low temperature promotes deterministic and more replicable outputs. 

For direct model evaluations on \textbf{GT Bench}, each model listed in \autoref{tab:all_models} was tested on 10,000 instances. For the evaluation of all baseline methods and our GTA framework, each method was tested on 1,000 instances from each respective dataset (GT Bench, GraCoRe, and NLGraph).

\subsection{Evaluation Scope and Statistical Protocol}
\label{app:evaluation_scope}

\textbf{Synthetic, domain-independent tasks.} GT Bench consists of synthetically generated instances of classical graph problems with algorithmically verified ground truth. This design isolates structural reasoning from application-specific knowledge. The transfer results on GraCoRe and NLGraph (\autoref{sec:gta_comparative_performance}) demonstrate gains beyond the training benchmark. The evaluation does not establish performance on domain-attributed graphs or tasks that require social, molecular, or other application-specific knowledge.

\textbf{Graph scale.} GT Bench instances range from 6 to 60 nodes, with per-family ranges set by the prompt-fit budget of \autoref{sec:graphbench_dataset}. Two constraints jointly fix it. First, a representation-controlled comparison is only meaningful while every instance fits the context window under all four formats, and the token cost of dense graphs grows as $O(V^2)$: a 40-node dense graph already carries up to 780 edges, which is why dense families use tighter node ranges. Second, difficulty is governed by the size of the search space rather than by the node count itself: a 40-node Hamiltonian instance spans a search space beyond $10^{47}$ orderings, and node ranges are calibrated inversely to algorithmic complexity, from 6--10 nodes for combinatorial-counting tasks such as Cycle Count and Spanning Tree Count up to 60 nodes for tree tasks with linear-time algorithms. The benchmark therefore offers a controlled difficulty gradient rather than a uniform cap, and the regime is demonstrably far from saturated: the best evaluated model reaches 75.1\% on GT-H, and model-averaged accuracy remains below 1\% under each of the four representations on dense Min-Cost Max-Flow. The scalability study in \autoref{sec:appendix_scalability} evaluates larger graphs produced by the same generation pipeline; the reported accuracy results remain specific to the evaluated size ranges.

\textbf{Statistical protocol.} Scores in the model-representation sweeps of \autoref{sec:bench_findings} come from a single inference run per configuration, decoded at temperature 0.01. These sweeps characterize representation effects across task--representation--model configurations, but do not quantify run-to-run uncertainty for individual cells. The targeted analyses in \autoref{sec:experiments}, including the component ablations, selector checks, and truncation statistics, are averaged over three runs with reported standard deviations. The accompanying repository provides generation and evaluation scripts.

\section{Experimental Results}
\label{appendix:experimental_results}
\subsection{Best-Performing Representation Per Task}
\label{sec:appendix_best_formats}

\autoref{tab:best_formats} lists the best-performing input representation for each task--graph-type combination in GT Bench, based on average accuracy across all evaluated models. These results informed the design of the Heuristic-Based Selector in GTA.

\begin{table}[ht]
\centering
\caption{Best-performing input representation format (Best Repr.) per task and graph type, based on average accuracy across all evaluated models (task-level averages in \autoref{sec:appendix_task_representation_accuracy_full}).}
\label{tab:best_formats}
\small
\begin{minipage}[t]{0.50\textwidth}
\centering
\begin{tabular}{ll}
\toprule
Task \& Graph Type & Best Repr. \\
\midrule
Biconnected Components (Dense) & AM \\
Biconnected Components (Sparse) & SL \\
Bipartite (Dense) & AM \\
Bipartite (Sparse) & NL \\
Bridge Count (Dense) & AL \\
Bridge Count (Sparse) & NL \\
Connectivity (Dense) & AM \\
Connectivity (Sparse) & AL \\
Cycle Count (Dense) & AL \\
Cycle Count (Sparse) & AM \\
Eulerian Circuit (Dense) & AM \\
Eulerian Circuit (Sparse) & SL \\
Eulerian Path (Dense) & AM \\
Eulerian Path (Sparse) & AM \\
Hamiltonian Circuit (Dense) & AL \\
Hamiltonian Circuit (Sparse) & AL \\
Hamiltonian Path (Dense) & AL \\
Hamiltonian Path (Sparse) & AL \\
Maximum Clique (Dense) & AM \\
Maximum Clique (Sparse) & NL \\
Maximum Flow (Dense) & AM \\
Maximum Flow (Sparse) & AL \\
\bottomrule
\end{tabular}
\end{minipage}%
\hfill
\begin{minipage}[t]{0.48\textwidth}
\centering
\begin{tabular}{ll}
\toprule
Task \& Graph Type & Best Repr. \\
\midrule
Maximum Independent Set (Dense) & AM \\
Maximum Independent Set (Sparse) & AM \\
Min Cost Max Flow (Dense) & AM \\
Min Cost Max Flow (Sparse) & AL \\
Minimum Cut (Dense) & AM \\
Minimum Cut (Sparse) & AL \\
Minimum Cycle (Dense) & AM \\
Minimum Cycle (Sparse) & NL \\
Minimum Spanning Tree (Dense) & AL \\
Minimum Spanning Tree (Sparse) & SL \\
Second MST (Dense) & SL \\
Second MST (Sparse) & SL \\
Shortest Path (Dense) & AL \\
Shortest Path (Sparse) & AL \\
Spanning Tree Count (Dense) & AM \\
Spanning Tree Count (Sparse) & AL \\
Tree Centroid (Tree) & SL \\
Tree Diameter (Tree) & SL \\
Tree LCA (Tree) & SL \\
Tree Max Independent Set (Tree) & SL \\
Triangle Count (Dense) & AM \\
Triangle Count (Sparse) & NL \\
\bottomrule
\end{tabular}
\end{minipage}
\end{table}

\subsection{Task-Specific Performance Across Input Representations}
\label{sec:appendix_task_representation_accuracy_full}

To further investigate the impact of input graph representation on LLM performance, this section details the average accuracy achieved by the evaluated models for 43 reported task--structure settings in GT Bench, broken down by the four input modalities: Natural Language (NL), Structured Language (SL), Adjacency Matrix (AM), and Adjacency List (AL). These results underpin the representation-sensitivity analysis of \autoref{subsec:representation_sensitivity} and the design of the heuristic selector in the GTA framework.

\autoref{tab:appendix_full_task_representation_results_ordered} tabulates the average accuracy for these task and graph type combinations, ordered alphabetically by task name. The accuracy figures represent the mean performance across all LLMs tested on that particular configuration. For each task--graph-type row, the highest accuracy is shown in bold.

\begin{table}[htbp]
\centering
\caption{Model-averaged accuracy (\%) for 43 task--structure settings in GT Bench under four input representations (NL, SL, AM, AL). Bold values indicate the best representation within each row.}
\label{tab:appendix_full_task_representation_results_ordered}
\resizebox{\textwidth}{!}{%
\begin{tabular}{lccccc}
\toprule
\textbf{Task \& Graph Type} & \textbf{NL (\%)} & \textbf{SL (\%)} & \textbf{AM (\%)} & \textbf{AL (\%)} & \textbf{Best Repr.} \\
\midrule
Biconnected Components (Dense) & 67.61 & 59.75 & \textbf{69.83} & 69.03 & AM \\
Biconnected Components (Sparse) & 17.85 & \textbf{23.31} & 18.90 & 22.98 & SL \\
Bipartite (Dense) & 95.63 & 95.81 & \textbf{97.64} & 96.45 & AM \\
Bipartite (Sparse) & \textbf{89.92} & 83.60 & 75.91 & 77.02 & NL \\
Bridge Count (Dense) & 83.01 & 71.93 & 84.55 & \textbf{88.23} & AL \\
Bridge Count (Sparse) & \textbf{42.91} & 36.55 & 35.14 & 40.88 & NL \\
Connectivity (Dense) & 88.79 & 89.01 & \textbf{90.65} & 87.89 & AM \\
Connectivity (Sparse) & 87.33 & 87.95 & 60.52 & \textbf{89.07} & AL \\
Cycle Count (Dense) & 12.67 & 10.98 & 11.25 & \textbf{13.01} & AL \\
Cycle Count (Sparse) & 29.78 & 24.03 & \textbf{31.84} & 26.35 & AM \\
Eulerian Circuit (Dense) & 83.05 & 75.01 & \textbf{95.03} & 92.99 & AM \\
Eulerian Circuit (Sparse) & 99.75 & \textbf{100.00} & 98.98 & 99.81 & SL \\
Eulerian Path (Dense) & 71.10 & 75.05 & \textbf{96.75} & 93.88 & AM \\
Eulerian Path (Sparse) & 98.99 & 98.15 & \textbf{100.00} & 96.42 & AM \\
Hamiltonian Circuit (Dense) & 81.35 & 80.99 & 83.03 & \textbf{88.27} & AL \\
Hamiltonian Circuit (Sparse) & 83.01 & 86.18 & 82.75 & \textbf{87.34} & AL \\
Hamiltonian Path (Dense) & 84.48 & 81.89 & 84.71 & \textbf{85.65} & AL \\
Hamiltonian Path (Sparse) & 67.68 & 68.51 & 69.03 & \textbf{71.97} & AL \\
Maximum Clique (Dense) & 24.65 & 22.08 & \textbf{38.61} & 34.87 & AM \\
Maximum Clique (Sparse) & \textbf{50.63} & 43.72 & 46.88 & 43.41 & NL \\
Maximum Flow (Dense) & 15.21 & 16.95 & \textbf{24.17} & 22.93 & AM \\
Maximum Flow (Sparse) & 34.88 & 28.03 & 33.45 & \textbf{39.56} & AL \\
Maximum Independent Set (Dense) & 27.19 & 32.61 & \textbf{36.93} & 33.17 & AM \\
Maximum Independent Set (Sparse) & 36.03 & 35.75 & \textbf{40.31} & 31.48 & AM \\
Min Cost Max Flow (Dense) & 0.13 & 0.00 & \textbf{0.98} & 0.07 & AM \\
Min Cost Max Flow (Sparse) & 21.51 & 21.20 & 23.01 & \textbf{23.45} & AL \\
Minimum Cut (Dense) & 14.70 & 15.22 & \textbf{19.03} & 18.65 & AM \\
Minimum Cut (Sparse) & 40.28 & 36.88 & 39.15 & \textbf{40.67} & AL \\
Minimum Cycle (Dense) & 86.99 & 83.07 & \textbf{87.35} & 85.33 & AM \\
Minimum Cycle (Sparse) & \textbf{81.42} & 72.49 & 80.98 & 75.03 & NL \\
Minimum Spanning Tree (Dense) & 14.39 & 16.92 & 17.22 & \textbf{19.81} & AL \\
Minimum Spanning Tree (Sparse) & 29.77 & \textbf{31.85} & 24.91 & 29.79 & SL \\
Second MST (Dense) & 14.68 & \textbf{19.83} & 11.79 & 18.09 & SL \\
Second MST (Sparse) & 23.77 & \textbf{24.15} & 22.96 & 23.70 & SL \\
Shortest Path (Dense) & 49.41 & 43.73 & 51.99 & \textbf{53.17} & AL \\
Shortest Path (Sparse) & 71.92 & 69.95 & 63.97 & \textbf{72.83} & AL \\
Spanning Tree Count (Dense) & 12.11 & 16.05 & \textbf{16.47} & 12.69 & AM \\
Spanning Tree Count (Sparse) & 18.66 & 18.08 & 19.50 & \textbf{19.85} & AL \\
Tree Centroid (Tree) & 56.25 & \textbf{60.84} & 42.59 & 51.09 & SL \\
Tree Diameter (Tree) & 37.73 & \textbf{42.05} & 34.01 & 36.03 & SL \\
Tree Max Independent Set (Tree) & 33.99 & \textbf{37.01} & 24.92 & 36.58 & SL \\
Triangle Count (Dense) & 7.82 & 16.08 & \textbf{23.29} & 18.99 & AM \\
Triangle Count (Sparse) & \textbf{40.33} & 39.17 & 37.45 & 37.77 & NL \\
\bottomrule
\end{tabular}%
}
\end{table}

The results in \autoref{tab:appendix_full_task_representation_results_ordered} show that no single input representation is universally optimal.

\textbf{Natural Language (NL)} achieves the highest average accuracy on sparse Bipartite Check, Bridge Count, Minimum Cycle, Maximum Clique, and Triangle Count.

\textbf{Structured Language (SL)} performs best on Tree Centroid, Tree Diameter, and Tree Max Independent Set. It also leads on sparse Minimum Spanning Tree and on Second MST at both densities.

\textbf{Adjacency Matrix (AM)} performs best on sparse Cycle Count, Eulerian Path at both densities, dense Maximum Clique, and Maximum Independent Set at both densities.

\textbf{Adjacency List (AL)} leads on sparse Connectivity, Hamiltonian Path and Circuit at both densities, and Shortest Path at both densities. It also achieves the highest average accuracy on sparse Maximum Flow, Min Cost Max Flow, and Minimum Cut.

The spread between the best and worst representation frequently exceeds ten points within a single row, and the winning format changes from row to row; this per-task granularity underlies the aggregated analysis in \autoref{sec:model_performance_gtbench}.

\section{Case Studies}
\label{sec:case_studies}

To further illustrate the practical implications of input representation sensitivity and motivate the adaptive approach of GTA, we present several case studies. These examples show cases where the same LLM produces different outcomes on the identical graph problem instance, based solely on the input format used to represent the graph structure.

\begin{figure}[htbp]
    \centering
    \includegraphics[width=\linewidth,height=0.90\textheight,keepaspectratio]{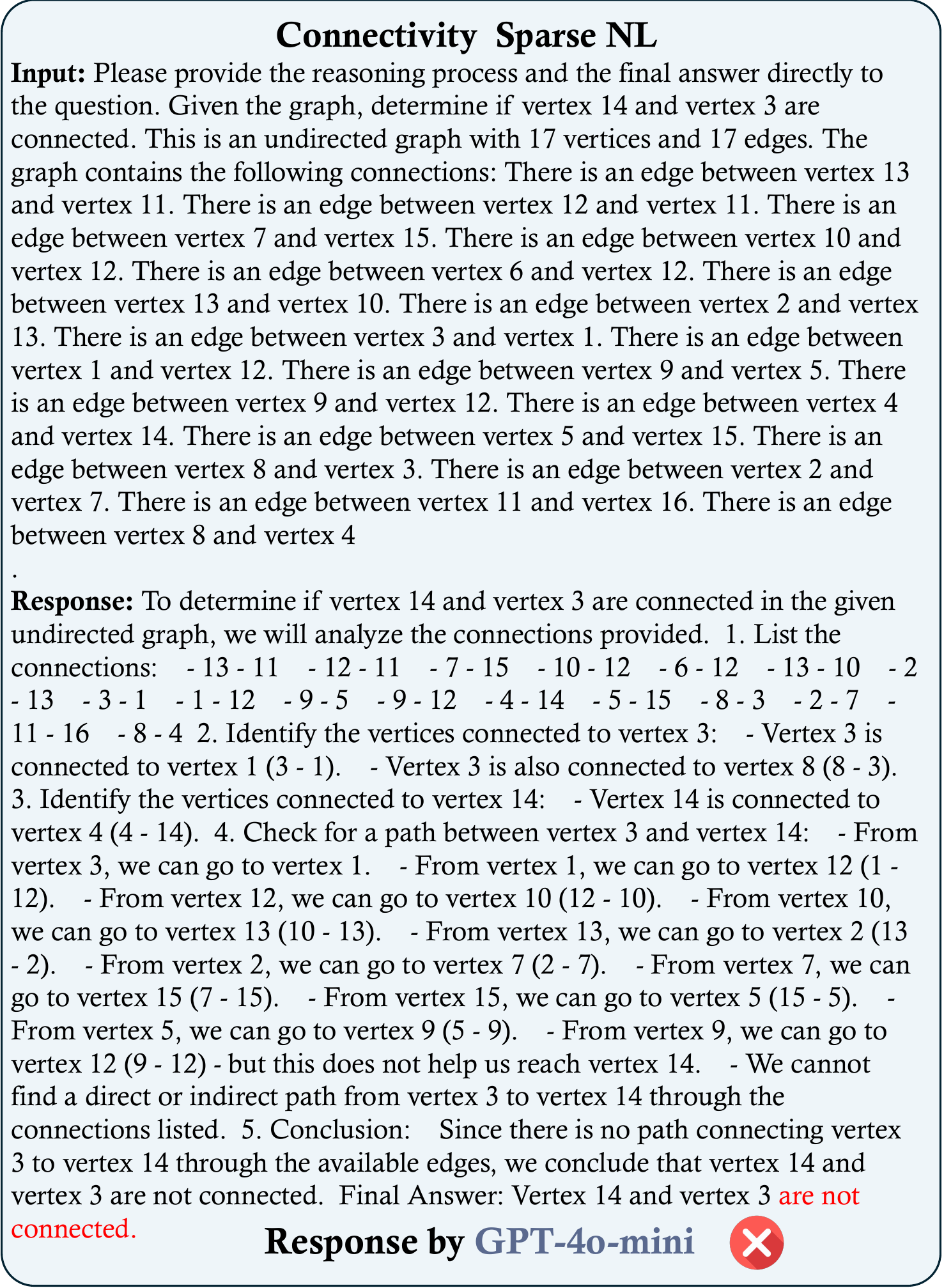}
    \caption{GPT-4o-mini incorrectly solves Connectivity (Sparse) using NL input.}
    \label{fig:case1}
\end{figure}

Our first comparison involves \texttt{GPT-4o-mini} tasked with determining connectivity on a sparse graph. As shown in \autoref{fig:case1}, when presented with the Natural Language (NL) description of the graph, the model incorrectly concludes the connectivity status between the specified nodes.

\begin{figure}[htbp]
    \centering
    \includegraphics[width=\linewidth,height=0.90\textheight,keepaspectratio]{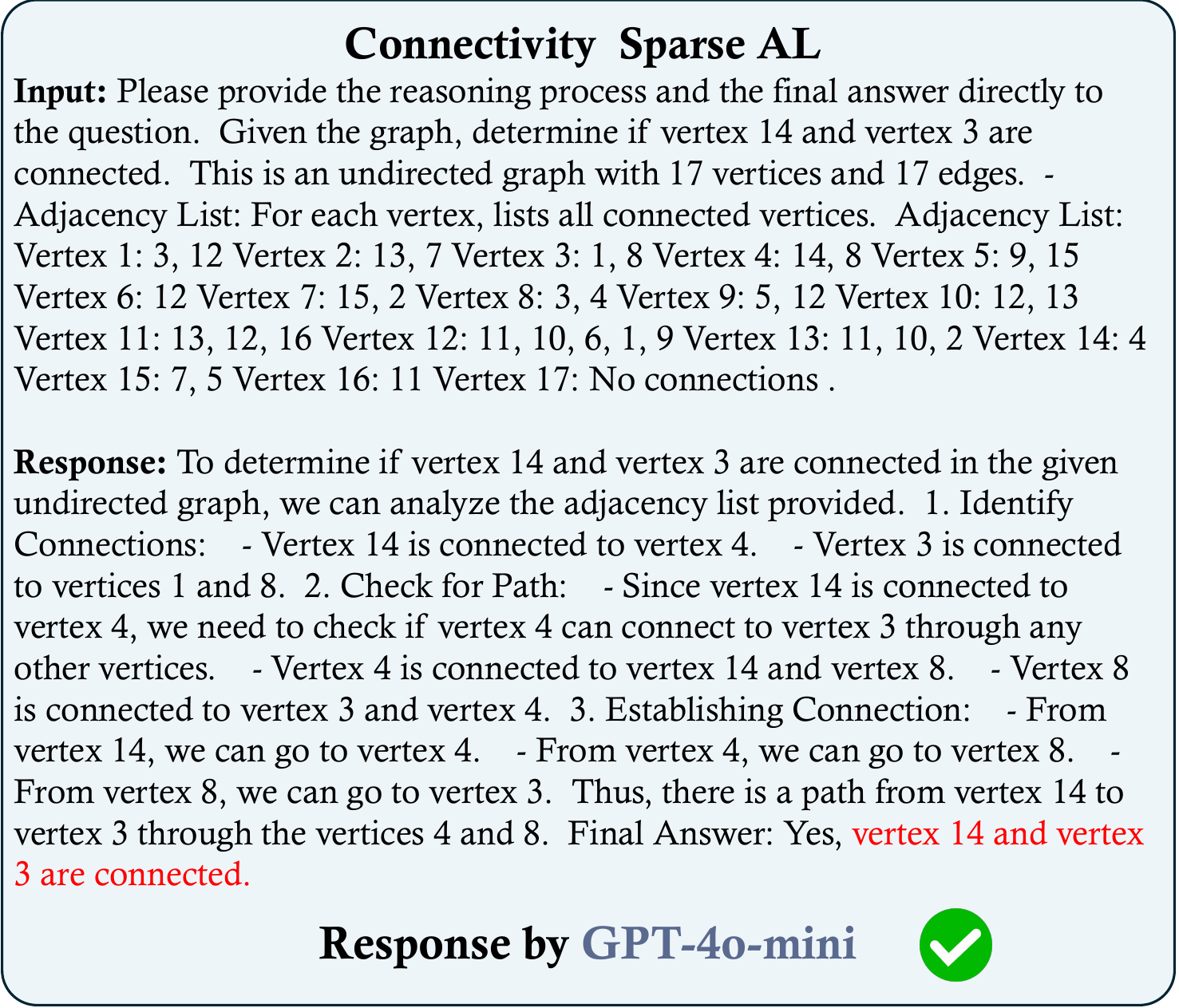}
    \caption{GPT-4o-mini correctly solves the same Connectivity (Sparse) problem using AL input.}
    \label{fig:case2}
\end{figure}

However, when the exact same graph instance and query are provided using the Adjacency List (AL) format (\autoref{fig:case2}), \texttt{GPT-4o-mini} successfully determines the correct connectivity. This suggests that the explicit listing of neighbors provided by AL facilitates the path-finding reasoning inherent in connectivity checks, which the model struggles to extract reliably from the more verbose NL description in this sparse setting.

\begin{figure}[htbp]
    \centering
    \includegraphics[width=\linewidth,height=0.90\textheight,keepaspectratio]{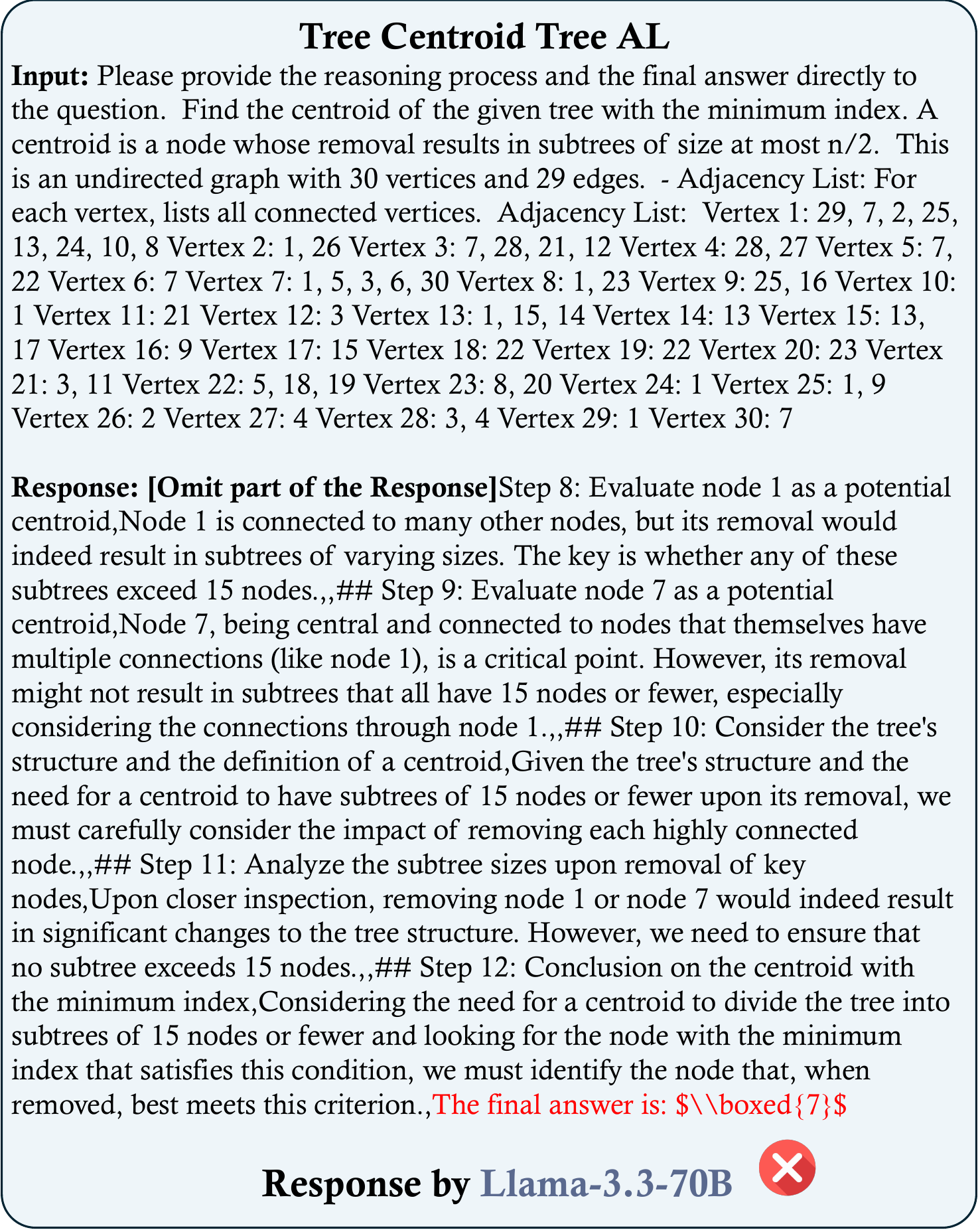}
    \caption{Llama-3.3-70B incorrectly solves Tree Centroid (Tree) using AL input.}
    \label{fig:case3}
\end{figure}

Next, we examine \texttt{Llama-3.3-70B} attempting to find the centroid of a tree structure. Using the AL representation (\autoref{fig:case3}), the model fails to identify the correct centroid node(s).

\begin{figure}[htbp]
    \centering
    \includegraphics[width=\linewidth,height=0.90\textheight,keepaspectratio]{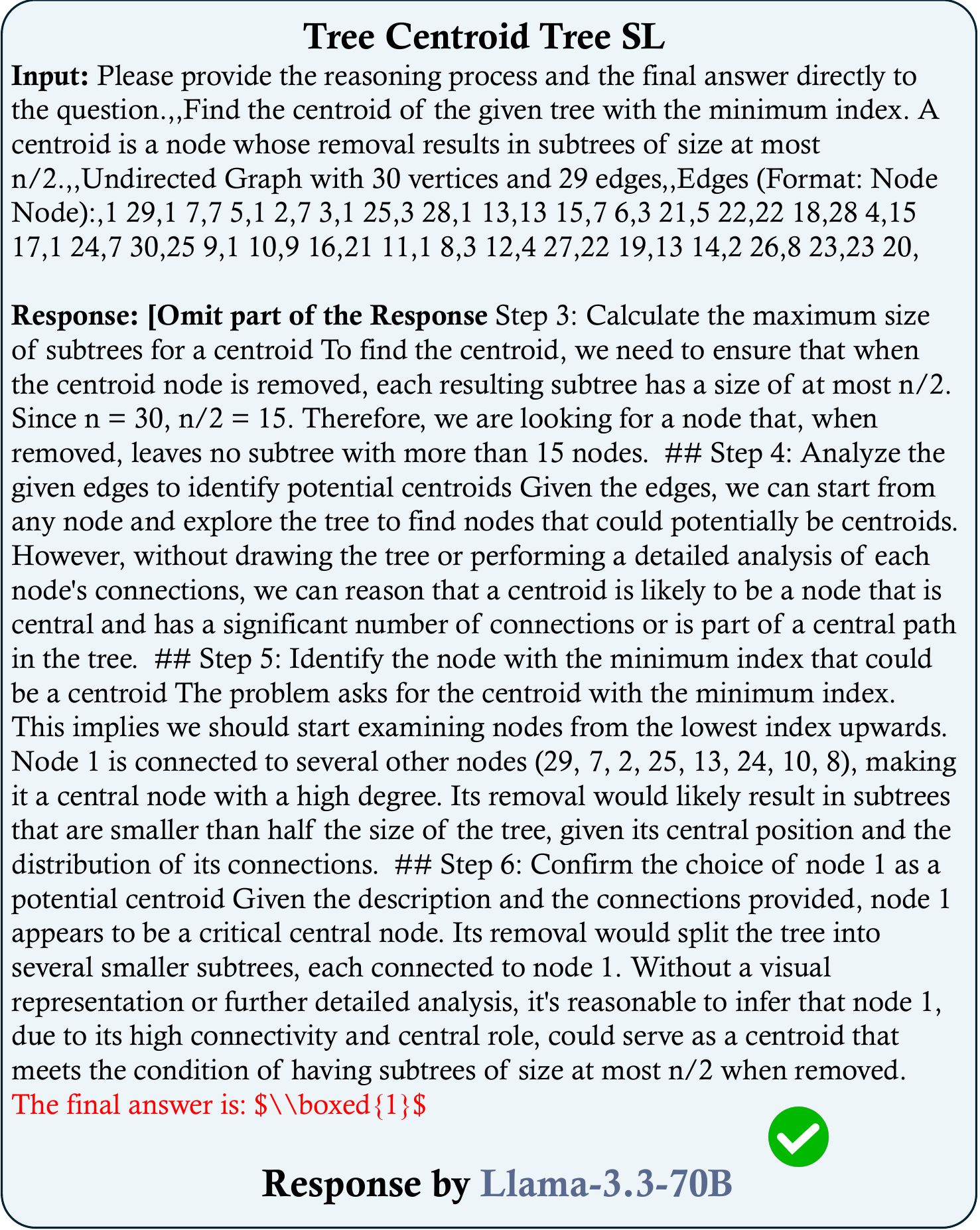}
    \caption{Llama-3.3-70B correctly solves the same Tree Centroid (Tree) problem using SL input.}
    \label{fig:case4}
\end{figure}

In contrast, presenting the tree structure via Structured Language (SL), as depicted in \autoref{fig:case4}, enables \texttt{Llama-3.3-70B} to accurately compute the tree centroid. The templated, potentially hierarchical nature of the SL format seems better suited for representing tree structures in a way that facilitates reasoning about subtree sizes and properties, which are crucial for the centroid calculation. The flat neighbor listing provided by AL appears less conducive to this type of structural reasoning for tree-specific algorithms.

\begin{figure}[htbp]
    \centering
    \includegraphics[width=\linewidth,height=0.90\textheight,keepaspectratio]{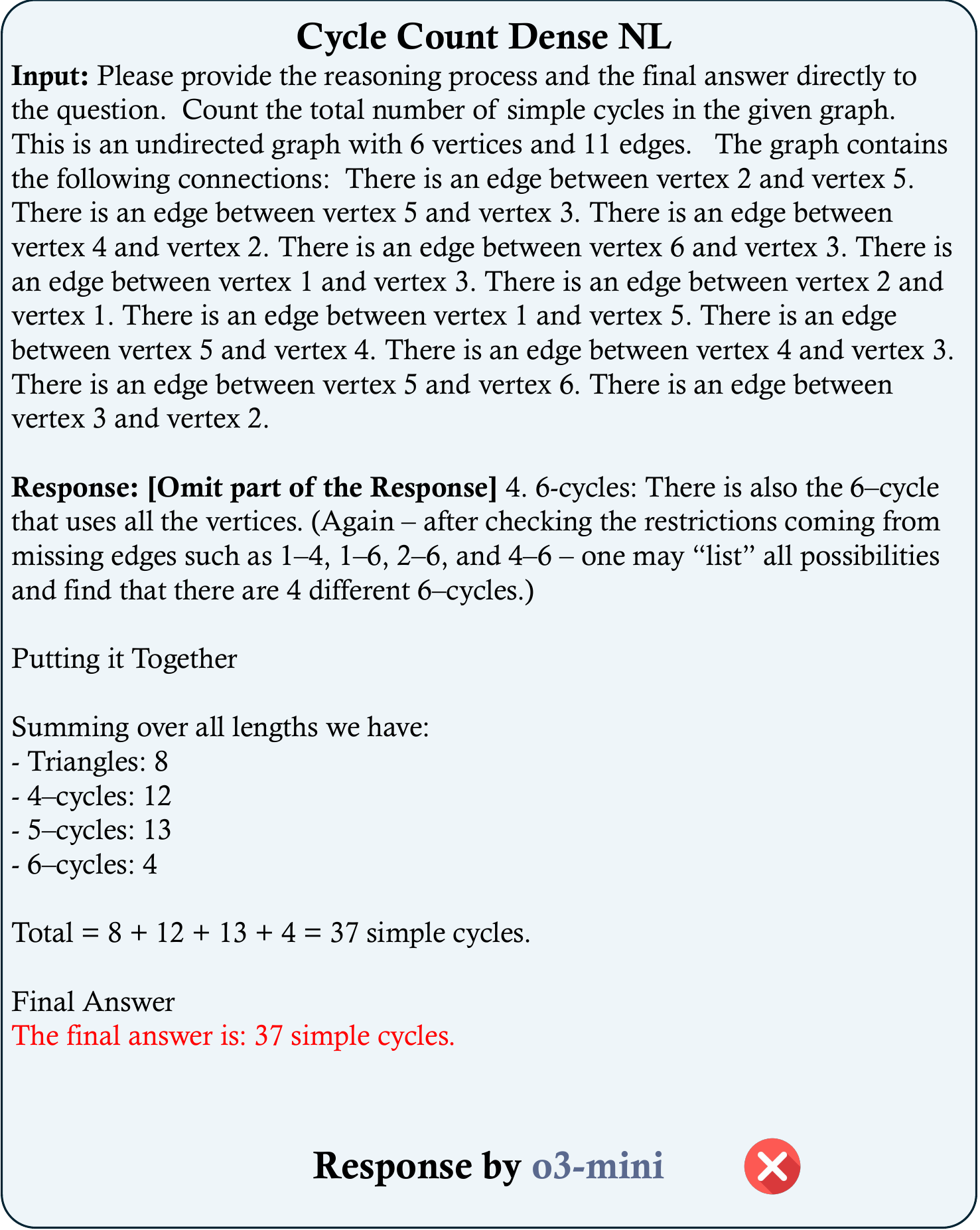}
    \caption{o3-mini incorrectly solves Cycle Count (Dense) using NL input.}
    \label{fig:case5}
\end{figure}

Our final case study features the highly capable \texttt{o3-mini} model tasked with counting cycles within a dense graph. When processing the NL description (\autoref{fig:case5}), the model makes an error in the cycle count.

\begin{figure}[htbp]
    \centering
    \includegraphics[width=\linewidth,height=0.90\textheight,keepaspectratio]{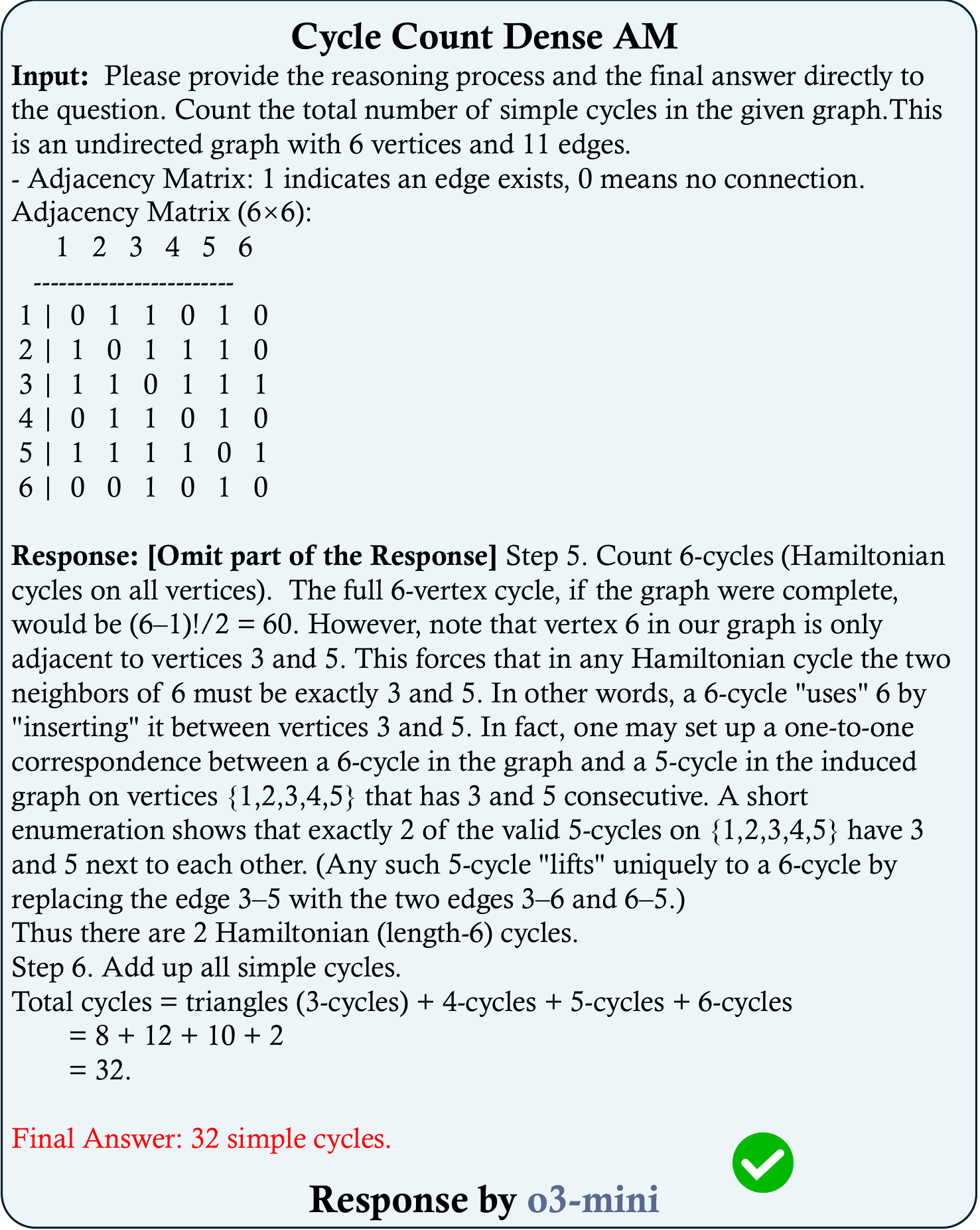}
    \caption{o3-mini correctly solves the same Cycle Count (Dense) problem using AM input.}
    \label{fig:case6}
\end{figure}

Yet, when provided with the Adjacency Matrix (AM) representation for the same dense graph (\autoref{fig:case6}), \texttt{o3-mini} accurately calculates the number of cycles. For dense graphs, where the number of potential edges is high, the AM offers a complete and explicit view of all connections and non-connections. Having every entry explicit appears to help the combinatorial bookkeeping that cycle counting requires, sidestepping the parsing ambiguities of the NL description.

In none of these cases is the failure a lack of capability in the abstract: the same model solves the same instance once the structure arrives in a form matched to the task. This is the instance-level counterpart of the aggregate sensitivity documented in \autoref{sec:bench_findings}, and it is precisely the behavior that GTA's Selector is trained to exploit.

\clearpage
\section{Prompt Templates}
\label{sec:prompt_templates}

This section details the prompt templates used for the core components of GTA: the Input Representation Selector, the Algorithm Generator, and the Algorithm Decomposer. Each prompt is designed to elicit a specific type of response from the LLM, guiding it through its designated sub-task.

\definecolor{academicblue}{RGB}{220, 230, 240} 
\definecolor{academicframe}{RGB}{100, 120, 140} 

\begin{figure}[ht]
\begin{tcolorbox}[colback=academicblue!30!white, colframe=academicframe, fontupper=\small, fonttitle=\bfseries, title={Prompt for Input Representation Selector}]
\textbf{Your Task: Optimal Graph Representation Selection}

You will be given a graph theory problem description. Your sole responsibility is to analyze this problem and select the \textbf{single best representation format} that would make it easiest for a system to process and solve the problem algorithmically.

\textbf{Available Representation Formats:}
\begin{itemize}
    \item \texttt{NL} (Natural Language: a descriptive text-based format)
    \item \texttt{SL} (Structured Language: a templated, keyword-based text format)
    \item \texttt{AM} (Adjacency Matrix: a matrix showing connections and/or weights)
    \item \texttt{AL} (Adjacency List: lists neighbors and/or weights for each vertex)
\end{itemize}

\textbf{Considerations for your selection:}
\begin{itemize}
    \item \textbf{Problem Type:} What kind of graph question is being asked (e.g., finding paths, checking properties, calculating flows)?
    \item \textbf{Graph Characteristics (infer from description):}
    \begin{itemize}
        \item Approximate size (number of nodes/edges).
        \item Is it likely sparse (few edges) or dense (many edges)?
        \item Are edge weights/capacities involved?
        \item Is it directed or undirected?
    \end{itemize}
    \item \textbf{Processing Efficiency:}
    \begin{itemize}
        \item \texttt{AM} can be good for dense graphs or when checking non-connections is important.
        \item \texttt{AL} is often efficient for sparse graphs and algorithms that traverse edges.
        \item \texttt{NL} and \texttt{SL} might be suitable for smaller graphs or problems where the structure is simpler to describe textually, but could be harder to parse for complex algorithms.
    \end{itemize}
\end{itemize}

\textbf{Input Problem:} \texttt{\{question\}}

\textbf{Instruction:}
Review the input problem. Based on your analysis and the considerations above, output \textbf{only one} of the following four representation format codes: \texttt{NL}, \texttt{SL}, \texttt{AM}, or \texttt{AL}.

\textbf{Your Output (must be one of NL, SL, AM, AL):}
\end{tcolorbox}
\caption{Prompt template for the Input Representation Selector component.}
\label{fig:prompt_selector}
\end{figure}

\begin{figure}[ht]
\begin{tcolorbox}[colback=academicblue!30!white, colframe=academicframe, fontupper=\small, fonttitle=\bfseries, title={Prompt for Algorithm Generator}]
\textbf{You are an Expert Algorithm Planner for Graph Theory Problems.}

\textbf{Your Task:}
You will be provided with a graph theory problem description. Your objective is to devise a \textbf{high-level strategic plan} or the \textbf{main algorithmic approach} that would be used to solve this problem.

\textbf{Crucial Instructions:}
\begin{itemize}
    \item \textbf{DO NOT solve the problem.}
    \item \textbf{DO NOT derive the final answer.}
    \item \textbf{DO NOT write code.}
\end{itemize}

Your output should be a \textbf{conceptual outline} of the algorithm or the logical phases involved. Think of it as describing the \textit{methodology} at a high level.

\textbf{For example, if the problem were "Find the shortest path between node A and node B":}
\begin{itemize}
    \item A good high-level plan might be: "Utilize a breadth-first search (BFS) starting from node A, keeping track of distances, until node B is reached. Alternatively, if edges have weights, apply Dijkstra's algorithm."
    \item A bad (too detailed/executing) response would be: "1. Initialize distance to A as 0, all others as infinity. 2. Add A to queue. 3. While queue not empty..."
\end{itemize}

\textbf{Considerations for your plan:}
\begin{itemize}
    \item Identify the core objective of the problem (e.g., finding a path, counting components, optimizing a value).
    \item What general class of graph algorithms is typically used for such a problem?
    \item What are the main conceptual stages of that algorithm?
\end{itemize}

\textbf{Input Problem Description:} \texttt{\{question\}}

\textbf{Instruction:}
Based on the problem description, provide a concise, high-level algorithmic plan or strategy. Focus on the "what" and "why" of the approach, not the detailed "how."

\textbf{Your High-Level Algorithmic Plan:}
\end{tcolorbox}
\caption{Prompt template for the Algorithm Generator component.}
\label{fig:prompt_generator}
\end{figure}

\begin{figure}[ht]
\begin{tcolorbox}[colback=academicblue!30!white, colframe=academicframe, fontupper=\small, fonttitle=\bfseries, title={Prompt for Algorithm Decomposer}]
\textbf{You are an Expert Algorithm Decomposer.}

\textbf{Your Task:}
You will be given an original graph theory problem description (\texttt{question}) and a high-level algorithmic plan (\texttt{plan}) designed to solve it. Your responsibility is to decompose this \texttt{plan} into a sequence of \textbf{2 or 3 highly detailed, specific, and actionable sub-steps}. These sub-steps should guide another system (an "Executor") to carry out the plan and solve the original \texttt{question}.

\textbf{Crucial Instructions:}
\begin{itemize}
    \item \textbf{Extreme Detail Required:} Each sub-step must be very specific. Think about what data structures are needed, what values to initialize, what to iterate over, what conditions to check, what information to maintain or update at each stage.
    \item \textbf{Actionable by an LLM:} Phrase each sub-step as a clear instruction that an LLM could follow to perform a part of the algorithm.
    \item \textbf{Logical Sequence:} The sub-steps must follow a logical order that reflects the progression of the \texttt{plan}.
    \item \textbf{Coverage:} Together, the sub-steps should comprehensively cover the entire \texttt{plan}.
    \item \textbf{DO NOT solve the original problem or provide the final answer.} You are only creating the detailed execution blueprint.
\end{itemize}

\textbf{Input:}

1.  \textbf{Original Problem (\texttt{question}):} \texttt{\{question\}}

2.  \textbf{High-Level Algorithmic Plan (\texttt{plan}):} \texttt{\{plan\}}

\textbf{Output Format (Strictly Adhere to This):}
You MUST output the decomposed sub-steps in the following exact format. Use \texttt{\#\#\# Sub-step X:} for each step.

\begin{lstlisting}[basicstyle=\ttfamily\footnotesize, frame=tb, breaklines=true, columns=flexible, xleftmargin=2em, rulecolor=\color{gray!60}]
### Sub-step 1:
[Provide an extremely detailed instruction for the first phase of the algorithm based on the plan. Be specific about:
- Initializing any necessary data structures (e.g., distance arrays, visited sets, queues, stacks, flow matrices).
- Setting initial values for variables or nodes (e.g., source node distance to 0, all others to infinity).
- The very first set of operations to perform.]
### Sub-step 2:
[Provide an extremely detailed instruction for the second phase, logically following Sub-step 1. Be specific about:
- Iterative processes (e.g., "While the queue is not empty..." or "For each neighbor of the current node...").
- Conditions for updates or state changes (e.g., "If a shorter path is found..." or "If the capacity is greater than zero...").
- How to update data structures or values based on operations.
- What to do if a certain condition is met (e.g., target node reached, no more augmenting paths found).]
### Sub-step 3:
[Provide an extremely detailed instruction for the third and final phase, if necessary to complete the plan. Be specific about:
- Continuation of iterative processes.
- Termination conditions for the algorithm.
- How to derive any intermediate results needed for the final answer (e.g., "Sum the capacities of edges in the minimum cut set" or "Backtrack from the target node to reconstruct the path").
- Final checks or operations before the overall algorithm concludes.]
\end{lstlisting}

\textbf{Instruction:}
Based on the provided \texttt{question} and \texttt{plan}, generate 2 or 3 highly detailed sub-steps strictly following the output format above.
\end{tcolorbox}
\caption{Prompt template for the Algorithm Decomposer component.}
\label{fig:prompt_decomposer}
\end{figure}

\end{document}